\documentclass[letterpaper, 10 pt, conference]{ieeeconf}  % Comment this line out if you need a4paper
\usepackage[noadjust]{cite}
\usepackage{graphicx}
\usepackage{amsmath}

\usepackage{listings}
\IEEEoverridecommandlockouts                              % This command is only needed if 
\title{\LARGE \bf

\textit{Morpheus}: An A-sized AUV with morphing fins and algorithms for agile maneuvering

}

\author{Supun Randeni$^{1,*}$, Michael Sacarny$^{1}$, Michael Benjamin$^{1}$ and Michael Triantafyllou$^{1}$% <-this % stops a space
\thanks{$^{1}$Department of Mechanical Engineering, Massachusetts Institute of Technology, Cambridge, MA, USA}  %
\thanks{$^{*}$supun@mit.edu}%
\thanks{This work was supported by Lockheed Martin Corporation}% <-this % stops a space
}

\begin{document}

\maketitle
\thispagestyle{plain}
\pagestyle{plain}

%%%%%%%%%%%%%%%%%%%%%%%%%%%%%%%%%%%%%%%%%%%%%%%%%%%%%%%%%%%%%%%%%%%%%%%%%%%%%%%%
\begin{abstract}

We designed and constructed an \textit{A-sized} base autonomous underwater vehicle (AUV), augmented with a stack of modular and extendable hardware and software, including autonomy, navigation, control and high fidelity simulation capabilities (\textit{A-size} stands for the standard sonobuoy form factor, with a maximum diameter of 124 mm). Subsequently, we extended this base vehicle with a novel tuna-inspired morphing fin payload module (referred to as the \textit{Morpheus} AUV), to achieve good directional stability and exceptional maneuverability; properties that are highly desirable for rigid hull AUVs, but are presently difficult to achieve because they impose contradictory requirements. The morphing fin payload allows the base AUV to dynamically change its stability-maneuverability qualities by using morphing fins, which can be deployed, deflected and retracted, as needed.
The base vehicle and \textit{Morpheus} AUV were both extensively field tested in-water in the Charles river, Massachusetts, USA; by conducting hundreds of hours of operations over a period of two years. The maneuvering capability of the \textit{Morpheus} AUV was evaluated with and without the use of morphing fins to quantify the performance improvement. The \textit{Morpheus} AUV was able to showcase an exceptional turning rate of around 25-35 deg s$^{-1}$. A maximum turn rate improvement of around 35\% - 50\% was gained through the use of morphing fins.

\end{abstract}

% ==============================================
\section{Introduction}
% ==============================================

The directional stability of autonomous underwater vehicles (AUVs) ensure the ability to maintain a steady course with minimal corrective control actions in the presence of disturbances \cite{eda1972directional,minorsky1922directional,maki2018fundamental}. The agility, or the maneuverability, of an AUV is the potential to make rapid maneuvers in heading and depth planes. The stability and agility have contradictory requirements; i.e. static or controlled surfaces located towards the stern of the vehicle (e.g. rudders, elevators, fixed fins, shrouds, etc.) increase the directional stability; however, they also adversely affect the maneuverability, reducing the ability to make rapid turns. This is because the increment in the stability index of a vehicle due to stern control surfaces is often larger than the turning moment it provides \cite{armo2000relationship,jones1973effect,bandyopadhyay2002maneuvering,ziaeefard2016novel,varyani2003effect,triantafyllou2020biomimetic,bettle2014validating,randeni2022bioinspired}. 

Both stability and maneuverability are desired features for AUVs \cite{phillips2010use}; therefore, AUVs in general are designed for a middle ground performance, partially compromising both stability and maneuverability. Improving both these features simultaneously was not possible for torpedo-shaped AUVs because they impose contradictory requirements. 
However, in our recent work \cite{randeni2022bioinspired,triantafyllou2020biomimetic}, we theoretically as well as experimentally showed that both stability and maneuverability can be improved by dynamically altering the directional stability, adopting the concept of bio-inspired morphing fins. 

\subsection{Designing an A-sized ``base'' AUV}

An AUV is a complex system with a number of co-related subsystems. These subsystems can be primarily divided into two categories: (1) the base layer, and (2) the specialized layer. The base layer consists of components that are essential for basic autonomous operations of the vehicle (i.e. the base vehicle); for example, underwater navigation, basic autonomy, low-level control, basic communication, and related essential sensing capabilities.   
In general, AUVs are employed to conduct specific task(s) and mission(s). The specialized layer includes additional hardware and software components that are vehicle and application specific, which are built on top of the base layer. This layer may include additional hardware interfaces and drivers, sensor processing algorithms, autonomy algorithms, etc. For instance, a vehicle designed to conduct side-scan sonar mapping operations, the specialized layer will include the sonar related hardware components and specialized software modules such as sensor drivers, on-the-fly data processing and recording  software, and potential autonomy algorithms for adaptive sampling and mapping.
%In the case of the \textit{Morpheus} AUV, the operations of the morphing fin module falls under the specialized layer, which includes the hardware related to the morphing fin mechanism and software components that trigger the adaptive operation modes of the fins. Thus, we are able to use the base vehicle integrated with other payloads for different applications if required. 

In this work, we designed and constructed an \textit{A-sized} base AUV, augmented with a stack of modular hardware and software, including navigation, autonomy, control and high fidelity software-in-the-loop (SITL) and hardware-in-the-loop (HITL) simulation capabilities (note -- \textit{A-size} stands for the standard sonobuoy \cite{holler2014evolution} form factor, with a maximum diameter of 124 mm and a length of around 0.9 m, ensuring the ability to launch from standard sonobuoy launchers onboard a wide array of fixed wing and rotary wing air crafts, surface ships and submarines \cite{ematt}). Subsequently, leveraged from our previous work \cite{randeni2022bioinspired}, we extended the base vehicle with a new tuna-inspired morphing fin payload module design, where the fins that can be deployed, deflected and retracted as required, augmenting the vehicle with capability to dynamically change its stability-maneuverability qualities. The designed base vehicle with morphing fin payload mechanism is named as the \textit{Morpheus} AUV.

\subsection{Designing a morphing fin payload module}

Aquatic animals that specialize in cruising, such as tunas, require a higher directional stability to minimize the control action needed during cruising. Hence, they have streamlined bodies that are relatively stiff, limiting their body flexing to the last 30\% of their length \cite{webb1984form}. However, their prey consists of smaller fish that have high body flexibility, and, hence, by employing significant body curvature, they can turn very rapidly \cite{fish2003aquatic}. 
As a result, 
%tunas can utilize retractable dorsal fins to dynamically change the directional stability to conduct rapid maneuvers at high speed \cite{li2021hydrodynamic}. 
tunas are capable of systematically changing the shape of their body fins to dynamically change the directional stability to conduct rapid maneuvers at high speed \cite{li2021hydrodynamic,triantafyllou2017tuna} -- when the forward located fins are retracted, their body becomes more directionally stable, gaining the ability to stably cruise at high speeds using a small amount of control energy. When they need to make a rapid turn, especially at high speeds, they deploy the dorsal fins whose mere presence destabilizes the body, increasing maneuverability. In addition, active control of the ventral fins provides additional turning moment, and smooth transients of forces and moments to obtain the precise level of directional body stability for the intended maneuver \cite{fish2017control}.

Inspired by tuna's adaptation mechanism, our previous work \cite{randeni2022bioinspired,triantafyllou2020biomimetic} demonstrated the ability to implement an engineered design of retractable fins for rigid-hull, torpedo-shaped AUVs in order to dynamically vary the stability-maneuverability indices.
While there is a number of recent studies and designs on developing biomimetic AUVs \cite{du2019design,matthews2021fin,han2020hydrodynamics,white2021tunabot}, as suggested by \cite{fish2020advantages} and \cite{phillips2012nature}, there is still a large gap between fish-like vehicle platforms and their corresponding aquatic animal; hence, torpedo-shaped AUVs are still superior to biomimetic AUVs in terms of speed and endurance.
In addition, many missions tasked to AUVs, such as seabed mapping, anti-submarine warfare, and surveillance, cannot be performed by biomimetic vehicles due to their unsteady large lateral motions, mechanical noise, and difficulty to design a sufficiently large payload bay while fulfilling soft-body bio-mimetic requirements. 
However, bio-mimetic AUVs are superior in terms of their maneuverability and agility, as compared to traditional torpedo-shaped vehicles \cite{fish2020advantages}. 
 Therefore, the intention of our work is not to develop a biomimetic AUV, but rather to create a bio-inspired vehicle by replicating the resultant hydrodynamic effects for a rigid-body engineered design, in order to enhance the vehicle's operational performance.

Through theoretical derivations, towing tank experiments and analytical simulations, \cite{triantafyllou2020biomimetic} investigated the ability to alter the stability and maneuvering qualities of self-propelled, rigid
hull AUVs by employing morphing fins. Our previous work \cite{randeni2022bioinspired} further extended this by investigating the variation of stability-maneuverability with different vehicle configuration and appendage designs. The evaluated vehicle configurations included: (1) the bare hull vehicle, (2) bare hull with different sizes of stern control surfaces, (3) different sizes of stern control surfaces combined with forward fins (4) different sizes of forward fins, and (5) different locations of the forward fins. This investigation was carried out by employing mathematical analysis, captive model tests and maneuvering simulations; validated with free swimming experiments. 
A 1-meter long bare hull AUV, retrofitted with different 3D-printed static appendages was used to investigate the variation of turning rate with free-swimming experiments.

In this paper, we present the design and construction of an A-sized base vehicle, including: (1) hull form and appendage configuration; (2) base vehicle hardware design, including the mechanical design and the construction of actuators, internal electronics and embedded computer system; (3) base vehicle software design, including underwater navigation, basic autonomy, low-level control, basic communication, and related essential sensing capabilities. Subsequently, leveraged from our previous work \cite{randeni2022bioinspired} in terms of hydrodynamic design, we develop an operational morphing fin payload module, and outfit it into our A-sized base vehicle; creating the \textit{Morpheus} AUV. We present the design and construction of the morphing fin payload, including: (1) theoretical aspects; (2) mechanical and hardware design; and (3) software algorithms for adaptive control of the morphing fins. We demonstrate both the base vehicle and \textit{Morpheus} AUV in-water, in the Charles river, Massachusetts, USA by conducting hundreds of hours of operations over a period of two years, evaluating the maneuvering capability of the vehicle with and without the use of morphing fins to quantify the performance improvement.

%The outline of the paper is as follows...

% ==============================================
\section{Base vehicle hardware design}
% ==============================================

\begin{figure}[h]
\centering
\includegraphics[trim=140 0 240 0, clip,  width=0.48 \textwidth]{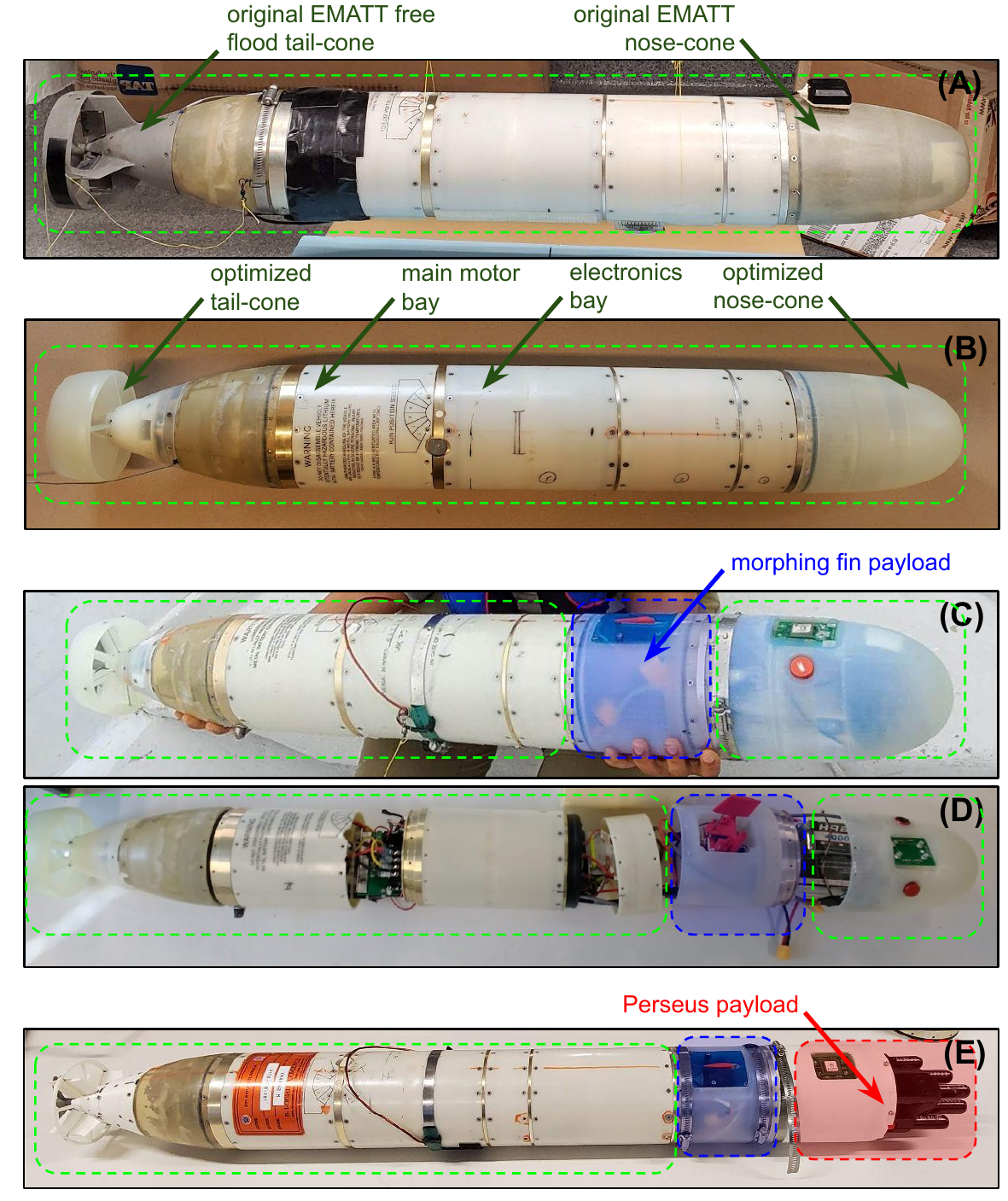}
\caption{The designed A-sized base AUV in four different configurations. Sections outlined in green, blue and red indicate base vehicle, morpheus payload and piUSBL payload components respectively. (A) The base vehicle in the conventional EMATT hullform; (B) base vehicle in the optimized hullform; (C) base vehicle appended with morphing fin payload (i.e. \textit{Morpheus} AUV); (D) \textit{Morpheus} AUV with disassembled hull sections; and (D) base vehicle appended with both morphing fin payload and piUSBL payload (i.e. \textit{Perseus} AUV)}
\label{fig:vehicle}
\end{figure}

The base vehicle developed in this work was a derivation of the expendable mobile anti-submarine warfare training target (EMATT) vehicle hullform, designed and produced by Lockheed Martin Corporation \cite{ematt}. Figure \ref{fig:vehicle}A shows the first iteration of our base vehicle, which utilized an EMATT shell, augmented with our own electronics and software stacks. The shell included the original EMATT nose-cone, empty body shell, main-motor bay that housed the original EMATT main motor, and a free-flood tail-cone with solenoid controlled control surfaces. Throughout this paper, we refer to this vehicle as the ``\textit{MIT-EMATT}''.

The second iteration of our base-vehicle; i.e the optimized edition, is shown in Figure \ref{fig:vehicle}B, which included hydrodynamically optimized nose and tail cones. The optimized nose-cone included an embedded GPS antenna, LED strobes, external pressure sensor and vacuum port; and the optimized tail-cone included four individually controlled, servo-driven control surfaces. These are further discussed in the following sections.  

Figures \ref{fig:vehicle}C and \ref{fig:vehicle}D show the base vehicle appended with the morphing fin payload module. When our base vehicle is appended with the morphing fin payload it is referred to as the \textit{Morpheus} AUV. The \textit{Morpheus} AUV had an overall length of 0.9 m and an \textit{A-sized} maximum diameter of 0.123 m. Figure \ref{fig:vehicle}E shows the base vehicle appended with morphing fin payload module as well as a piUSBL payload module \cite{rypkema2018closed,rypkema2021synchronous}, which is referred to as the \textit{Perseus} AUV. 

\subsection{Nose-cone design} \label{sec:nosecone}

\begin{figure}[h]
\centering
\includegraphics[trim=160 0 160 20, clip,  width=0.48 \textwidth]{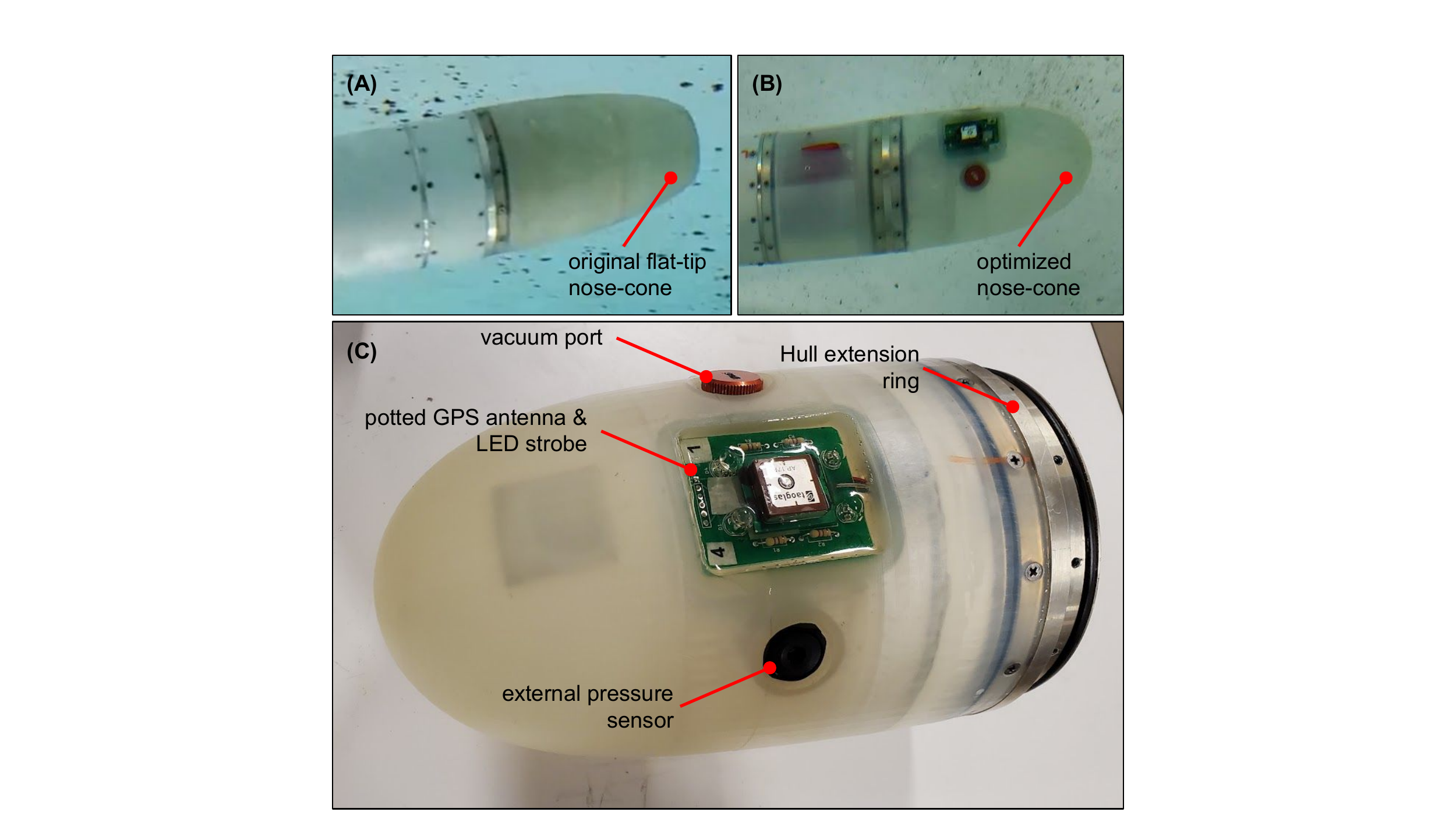}
\caption{Comparison between the (A) flat-tip nose-cone of the original EMATT platform, and (B) hydrodynamically optimized nose-cone. (C) The 3D printed optimized nose-cone consisted of two cylindrical slots to install the external pressure sensor and vacuum port; and a third rectangular slot to insert a circuit board containing the GPS antenna and LED strobes, which was potted with clear epoxy to make it waterproof.}
\label{fig:nosecone}
\end{figure}

The original EMATT platform had a nose-cone with a flat-tip, primarily to ensure a higher usable space density, which resulted in a higher drag coefficient (see Figure \ref{fig:nosecone}A). In this work, a new nose-cone was designed as shown in Figures \ref{fig:nosecone}B and \ref{fig:nosecone}C, which was optimized to reduce the hydrodynamic
resistance of the body as well as to preserve the usable space density inside the nose-cone \cite{mastersthesis}.

The optimized nose-cone was manufactured by 3D printing; which was connected to the vehicle body using a metal hull extending ring, as shown in Figure \ref{fig:nosecone}C. The nose-cone was designed with two cylindrical slots, which were used to install the vehicle's external pressure sensor sensor (i.e. Blue Robotics depth sensor \cite{bluerobotics_30bar} housing an MS5837-30BA pressure sensor \cite{MS5837_30BA}) to measure the vehicle depth, and a vacuum port. A third rectangular slot was also designed, which allowed to insert a circuit board containing the GPS antenna and LED strobe. Upon installation of the circuit board, the slot was potted with clear epoxy to ensure the water-tightness. The vehicle's main battery bank was housed inside the nose-cone.

\subsection{Tail-cone design} \label{sec:tailcone}

The original EMATT platform had a tail-cone with a solenoid-controlled single rudder and a single elevator. Due to solenoid control, they both were limited to three positions: hard-to-port, hard-to-starboard and neutral. In the \textit{MIT-EMATT} base-vehicle, we maintained the same actuator mechanism, connected to our own electronics and software.

\begin{figure}[h]
\centering
\includegraphics[trim=100 30 110 0, clip,  width=0.48 \textwidth]{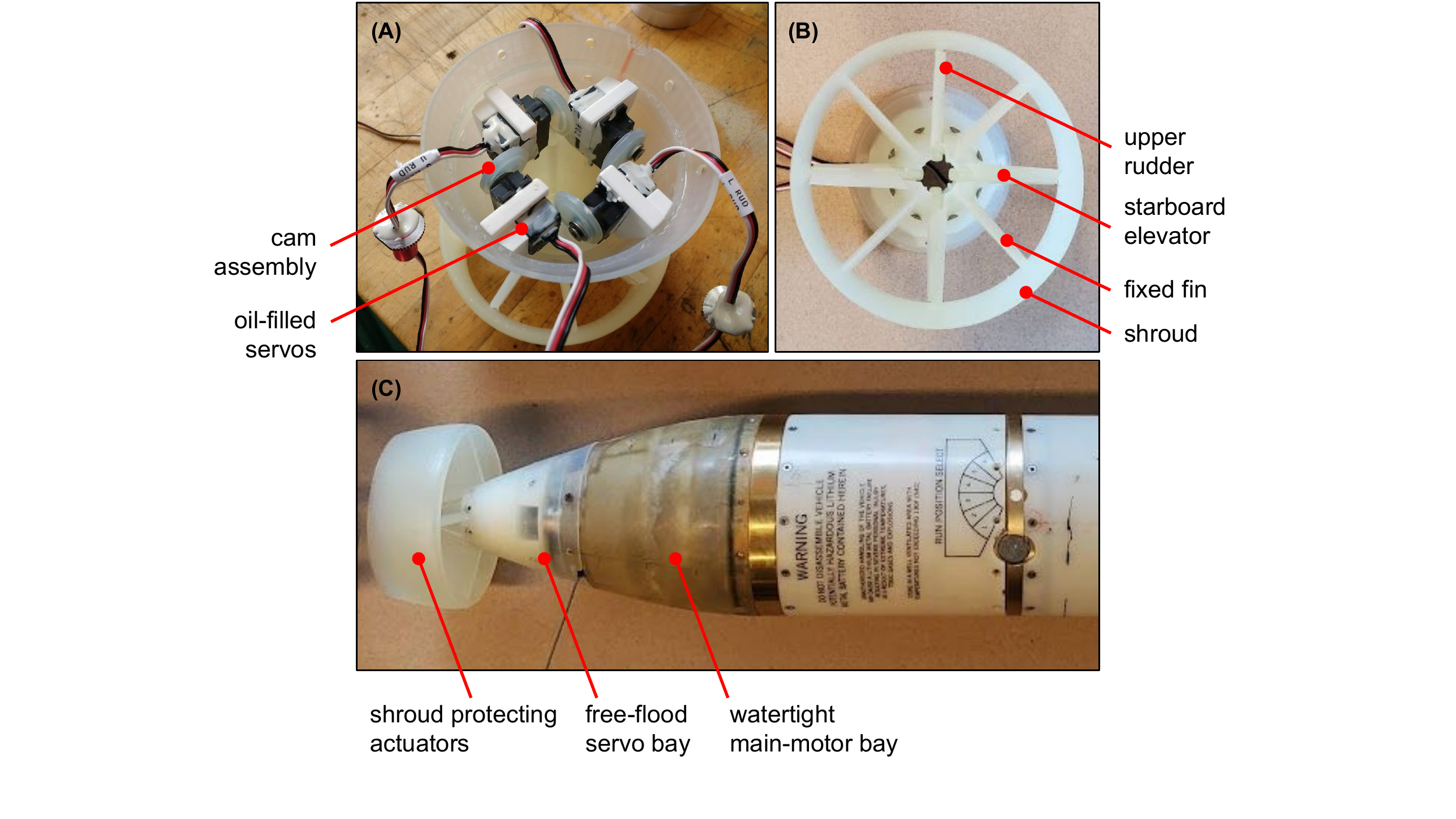}
\caption{The hydrodynamically optimized tail-cone with four independently controlled, servo-driven control surfaces. (A) The tail-cone assembly consisted of the four oil-filled micro-servos, cam and push-rod assemblies and control linkages, all mounted in a 3D-printed shell. (B) The two rudders, two elevators and propeller were protected by a shroud. The shroud was connected to the hull of the vehicle using four fixed fins, which had a fixed 3$^{\circ}$ deflection angle to counteract the propeller torque. (C) The free-flood, 3D-printed tail-cone section was separated from the watertight main-motor bay using a bulkhead, which had a shaft seal to penetrate the propeller drive shaft.}
\label{fig:tailcone_drive}
\end{figure}

The hydrodynamically optimized tail-cone version, as shown in Figure \ref{fig:tailcone_drive} had four cruciform-shaped, independently controlled, servo-driven control surfaces that can provide heading, pitch and roll control to the vehicle. Due to servo control, each control surface had the ability to be precisely controlled with a maximum articulation angle limit of around 15 degrees. The control surfaces and the propeller were protected by a shroud, which was an operational requirement. The shroud was connected to the hull of the vehicle using four fixed fins, which were not only acting as supports for the shroud, but their fixed 3$^{\circ}$ deflection angle also developed lift forces and hence a moment that counteracts the propeller torque. 

As shown in Figure \ref{fig:tailcone_drive}A, the tail-cone assembly consisted of the four control surfaces, their control linkages, and four servos, all mounted in a 3D-printed shell. The servos were held in place in the shell by 3D-printed dogs. Servo shaft rotation was converted to push-rod reciprocating action by cam assemblies. The push-rods then drove control surface articulation through control arms, while the control surfaces rotate around fixed pins.
Since the tail-cone module was free-to-flood and the utilized micro-servos were not intended for submersion, they were oil-filled in-house. The servo cables were transitioned to the watertight main-motor bay through marine epoxy-filled bulkhead penetrators. 

\subsection{Electronics design}

\begin{figure}[h]
\centering
\includegraphics[trim=0 380 40 100, clip,  width=0.48 \textwidth]{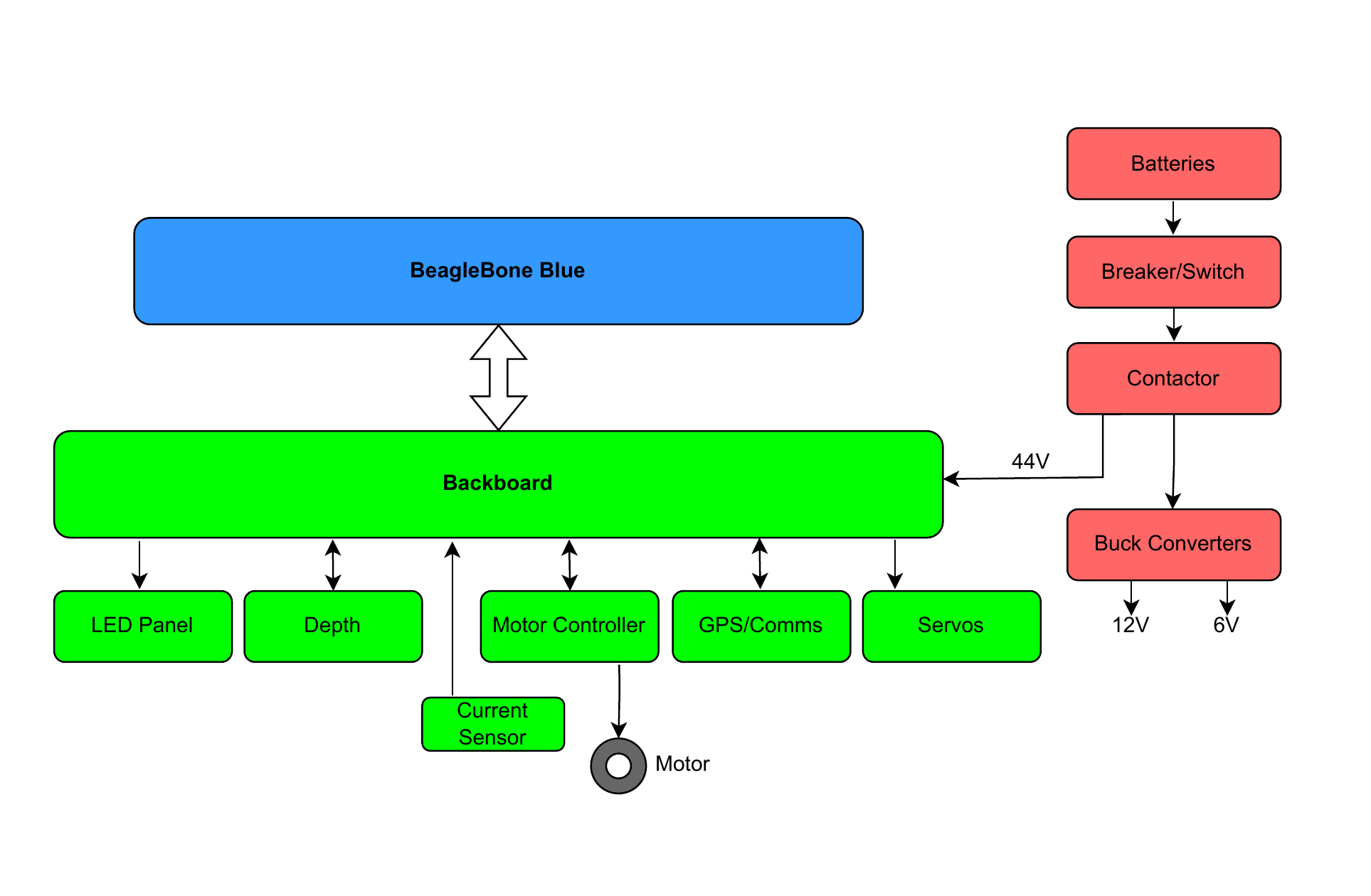}
\caption{A high-level overview of the electronics of the \textit{Morpheus} AUV.}
\label{fig:Morpheus_electronics}
\end{figure}

Figure \ref{fig:Morpheus_electronics} represents the overall electronics design. Off-the-shelf electronic components, including a BeagleBone Blue single-board computer and motor controller were physically and electrically joined on a central custom backboard printed circuit board (PCB). Wiring then connected the backboard to batteries, power controls, sensors, servos, and a custom LED strobe panel. 

\subsubsection{BeagleBone Blue} \hfill

The BeagleBone Blue single board computer \cite{grimmett2013beaglebone} was selected as the main vehicle computer. It contains an ARM processor \cite{furber1987acorn} that runs Linux operating system, as well as two Programmable Real-Time Units (PRUs), which are independent of the ARM processor; therefore, capable of quickly responding to inputs and produce very precisely timed outputs, such as PWM motor control outputs, similar to a micro-controller.
Thus, both low-level control as well as higher-level processes were able to be run on a single computer, reducing the complexity and saving physical space that is precious for A-sized AUVs. The BeagleBone Blue included analog to digital converters (ADC), general purpose input and output (GPIO), PWM support, embedded inertial measurement unit (IMU), I$^2$C interface, embedded WiFi, and universal asynchronous receiver-transmitter (UART) serial buses.

\subsubsection{Backboard} \hfill

A custom PCB served as an electronic and physical integration foundation for the BeagleBone and other components, including the motor controller, current sensor, GPS module, 6V power supply, power conductors, servo connections, and system cabling. Discrete components provided ADC conditioning and logic translation for the LED strobe.

\subsubsection{Motor controller} \hfill

The utilized main motor controller was a Cytron MD25HV 7-58V controller, selected to provide power in excess of 1 kW, should this be required for future designs, all in a compact and durable package. The standard EMATT motor required 150 W.

\subsubsection{Current sensor} \hfill

A Pololu 4046 current sensor produced an analog signal that was used to monitor motor current.

\subsubsection{Depth and barometric pressure sensors} \hfill

A Blue Robotics Bar30 depth sensor provided real-time depth readings via the I$^2$C bus to the BeagleBone. The BeagleBone itself has an ambient pressure sensor, useful for pre- and mid-test internal vacuum monitoring. 

\subsubsection{LED strobe} \hfill

The LED strobe panel was located in the nose-cone. It consisted of LEDs and a GPS antenna mounted on a custom PCB, all potted in clear epoxy.

\subsubsection{Servo drive} \hfill

The six micro-servos in the \textit{Morpheus} vehicle design were driven by conventional PWM signaling from the BeagleBone through the backboard.

\subsubsection{GPS and cellular modem} \hfill

GPS and cellular modem capabilities were provided by an Adafruit FONA 3G Cellular Breakout board. The antenna for this was routed to the nose-cone.

\subsubsection{Power supply} \hfill

The nominal battery voltage for the system was 44.4 volts, supplied by two 6 cell lithium-polymer (LiPo) batteries in series. These were connected to the rest of the system by a 30A combination circuit breaker/power switch, followed by a relay-operated DC contactor. The contactor was only activated when an external plug is inserted into a through-hull connector. This allowed us to control vehicle power after the hull has been closed and pressurized.

Raw battery voltage was brought down to 12V by an automotive-style buck converter and distributed to the BeagleBone and LED panel. A Pololu 4092 buck converter provided regulated 6V DC power for the Adafruit FONA GPS and 3G cellular module.

% ==============================================
\section{Base vehicle software design}
% ==============================================

In this work, we subdivided the critical components (both software and hardware) of the \textit{Morpheus} AUV into two categories; the base vehicle components, and specialized components. This modular subdivision allowed us to rationally reconfigure and re-purpose our A-sized base vehicles for other applications. 

\begin{figure}[h]
\centering
\includegraphics[trim=30 440 60 120, clip,  width=0.48 \textwidth]{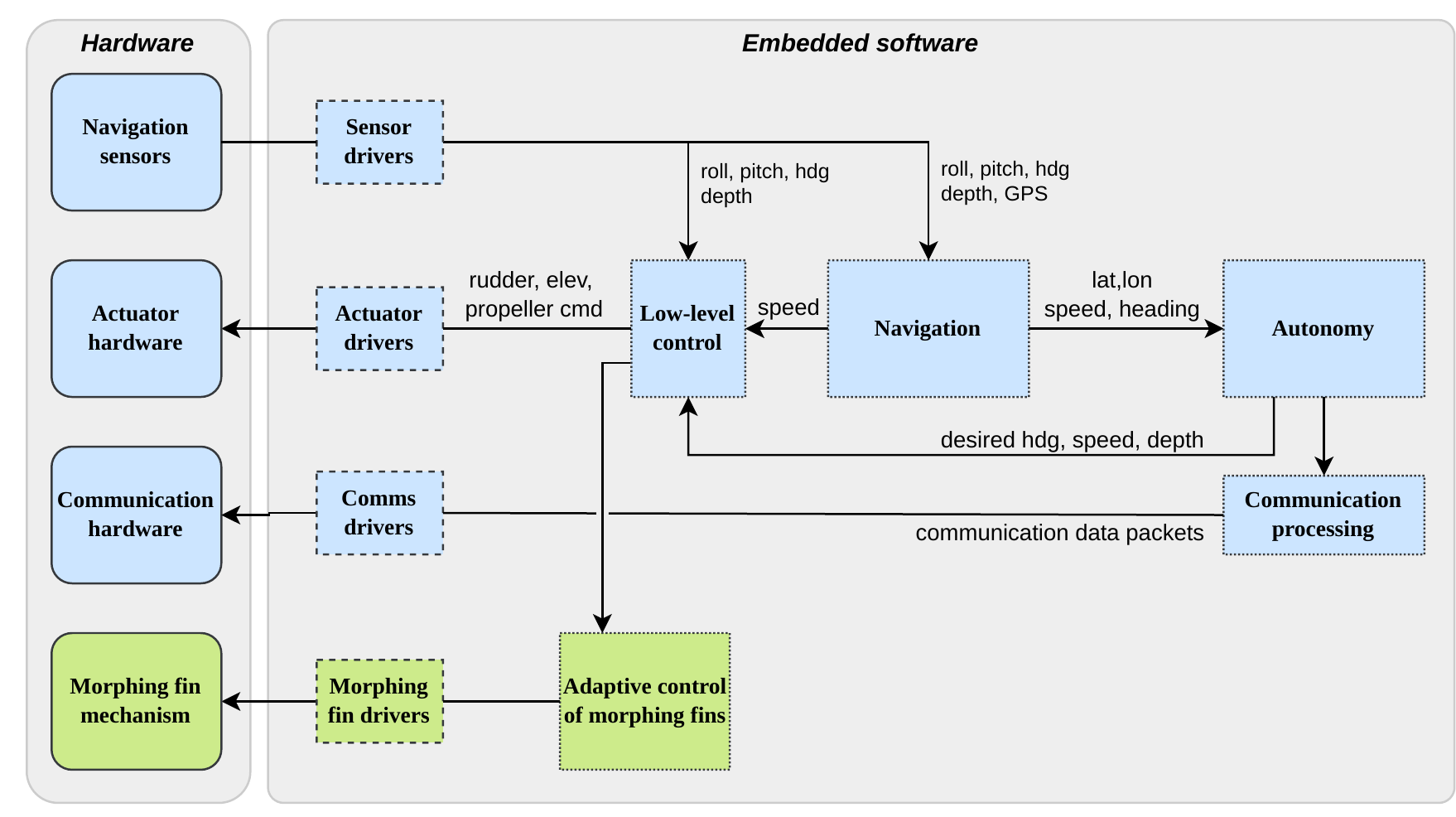}
\caption{A high-level overview of the hardware and software components of the \textit{Morpheus} AUV, together with the data flow among them. The components that belong to the base layer are filled in blue, while those belong to the specialized layer (i.e. the components related to morphing fins) are filled in green.}
\label{fig:system_overview}
\end{figure}

Figure \ref{fig:system_overview} illustrates a higher level overview of hardware and software components of the \textit{Morpheus} AUV, together with the information flow among them. The components that belong to the base layer are filled in blue, while those belong to the specialized layer (i.e. related to morphing fins, in this case) are filled in green. 
%The base layer consists of components that are essential for basic autonomous functions of the vehicle, including hardware and software components required for underwater navigation, basic autonomy, low-level control, basic communication, and related essential sensing capabilities of the vehicle.
The remainder of this section will discuss the vehicle software components in the base layer. 

% \subsection{Software architecture of the base AUV}
% ----------------------------------------------

The base software layer, shown in Figure \ref{fig:system_overview} with blue filled blocks, is responsible for all the essential functionalities of a vehicle that transforms the hardware package into an autonomous platform. This includes navigation, autonomy, low-level control, communication and interfacing with related sensor and actuator hardware.

% \subsubsection{Sensor-to-actuator information flow}

As seen from Figure \ref{fig:system_overview}, the sensor-to-actuator data flow begins with navigation sensor drivers, which communicate with external navigation sensor hardware, and provide raw sensor data to the vehicle's navigation system. The navigation software is responsible for estimating the vehicle's position (i.e. the latitude, longitude of the vehicle where it thinks it is at), depth, speed and attitude (roll, pitch and heading angles), while on the surface as well as underwater. The autonomy software is responsible for guiding the vehicle to the intended target location while avoiding no-go zones and obstacles in order to accomplish the mission tasks. The autonomous helm typically digests the navigation solution, and continually produces decisions on the desired heading, desired speed and desired depth commands that the vehicle needs to follow in order to achieve mission objectives. The low-level control software is responsible for executing the desired heading, speed and depth commands instructed by the autonomy system. This is achieved by controlling the actuators of the vehicle (e.g. propeller speed and control surface angle of attacks) to maintain the desired commands to the best of its ability. The actuator drivers then communicate these commands to the external actuator hardware; for example, using PWM, GPIO, and controller area network (CAN) bus commands.

With respect to our base vehicle software design, the components shown in Figure \ref{fig:system_overview} are subdivided into three modular sub-systems. The primary software hub, which is referred to as the \textit{MITFrontseat}, includes low-level sensor and actuator drivers, low-level control system, higher-level mission and safety management processes. The \textit{MITFrontseat} outsources the navigation task to a specialized navigation engine -- named \textit{HydroMAN 2.0}, which provides the vehicle's position estimate (i.e. navigation solution) to \textit{MITFrontseat}. The vehicle autonomy is either handled within \textit{MITFrontseat}, or can also be outsourced to a user's own payload autonomy software system. Each sub-component outlined in this paragraph are further explained in the following sections.

\subsection{Middlewares -- inter-process and inter-system communication}

As visualized in Figure \ref{fig:system_overview}, the base vehicle software is composed of a number of distributed components; e.g. low-level interfaces, navigation, autonomy and low-level control modules. Hence, using a suitable middleware, or a combination of several middlewares, to glue the software components together is important \cite{astley2001middleware,mohamed2008middleware}. There are a few different choices of middleware typically used by the marine robotics community, such as, common object request broker architecture (CORBA) \cite{watson1994omg}, mission oriented operating suite (MOOS) \cite{newman2008moos}, data distribution service (DDS) \cite{pardo2003omg},
 robotics operating system (ROS) \cite{quigley2009ros}, Goby3 \cite{schneider2016goby3}, lightweight communications and marshalling (LCM) \cite{huang2010lcm}, etc. In this work, we primarily use MOOS as the middleware for inter-process communication within the software sub-systems. In addition, a standardized interface that exchanges pre-defined, encoded google protocol buffer (protobuf) \cite{protobuf_guide} messages over the TCP communication architecture is also used for inter-system communications; for example, between \textit{MITFrontseat} and \textit{HydroMAN 2.0}, and between \textit{MITFrontseat} and payloads. This architecture was chosen to ensure the independence of each sub-system from others, and the ability to re-use these sub-systems in different frameworks, even if they use non-MOOS middleware. This is further discussed in Sections \ref{sec:hydroman_gateway} and \ref{sec:payload_interface}.

\subsection{Embedded computing system}

\begin{figure*}[h]
\centering
\includegraphics[trim=0 0 0 0, clip,  width=1.0 \textwidth]{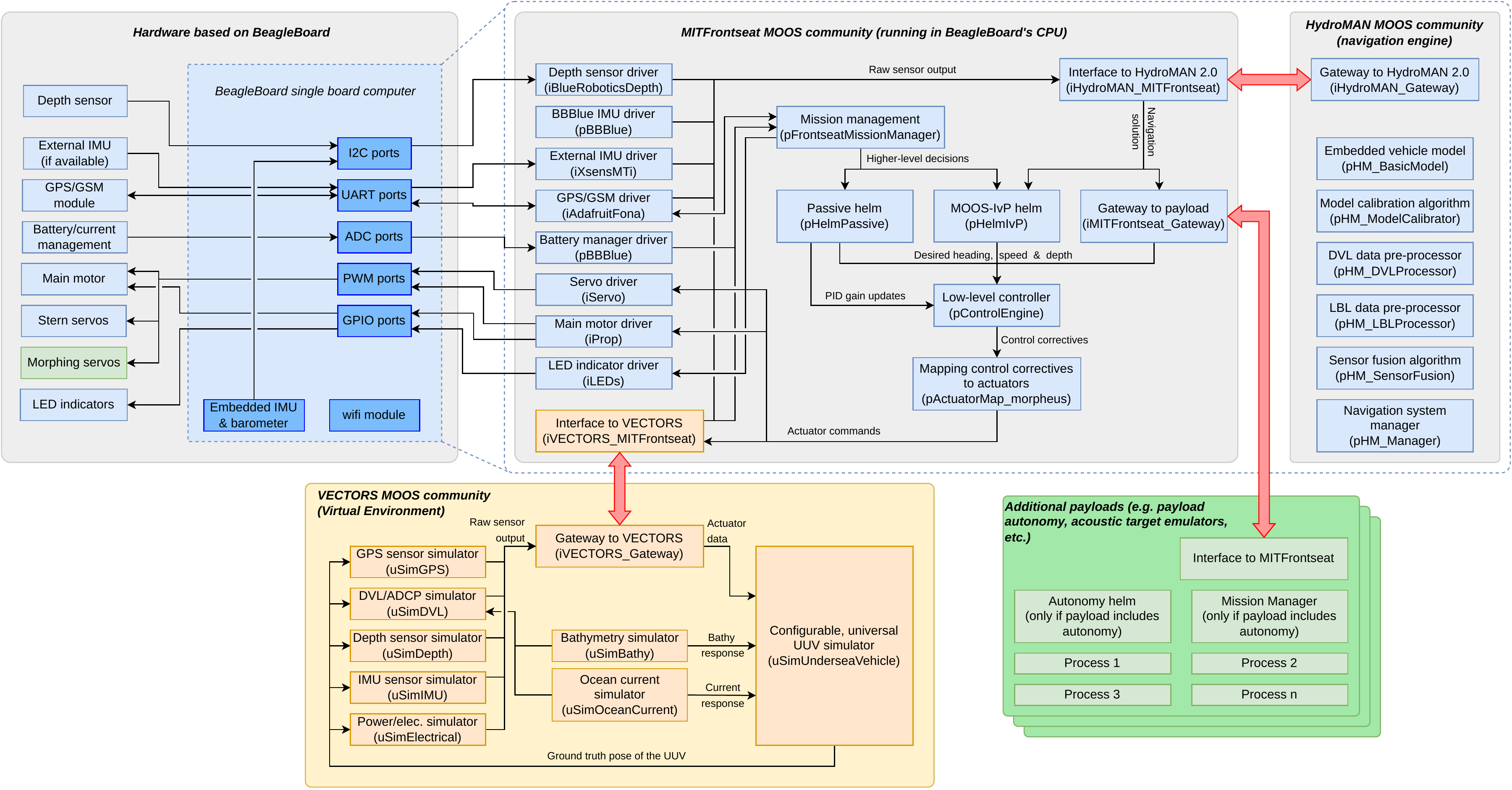}
\caption{A complete}
\label{fig:complete_system}
\end{figure*}

Figure \ref{fig:complete_system} illustrates the complete base-vehicle software diagram of the \textit{Morpheus} AUV. 
In this work, we developed a boilerplate frontseat software stack, referred to as the \textit{MITFrontseat}, that is sufficiently modular and generic to be utilized as a base-vehicle frontseat software for other types of AUV designs as well. As seen from Figure \ref{fig:complete_system}, the \textit{MITFrontseat} stack handles both low-level routines such driving low-level hardware (i.e. driving and communicating with sensors and actuators), while also handling higher-level processes such as navigation, autonomy, control and vehicle safety management.
Typically, a micro-controller is used to handle low-level routines, which is interfaced with a single board computer that runs higher level processes. 
In this work, however, we have used a BeagleBoard single board computer \cite{grimmett2013beaglebone} as the main vehicle computer. It contains an ARM processor \cite{furber1987acorn} that runs Linux operating system, as well as two PRUs, which are independent of the ARM processor; therefore, are capable of quickly responding to inputs and produce very precisely timed outputs, such as PWM motor control outputs, similar to a micro-controller.
Thus, both low-level routines as well as higher-level processes were able to be run on a single computer, reducing the complexity and saving physical space that is precious for A-sized AUVs. 

Since the main vehicle computer also handles low-level hardware, the {\textit MITFrontseat} included an array of MOOS drivers for various sensors and actuators that are typically used in micro AUVs. One of the drawbacks of this architecture is that some of these MOOS drivers are only supported for BeagleBoard computers; hence, if one anticipates to run {\textit MITFrontseat} on a different single board computer board, these low-level drivers may required to be modified accordingly. That said, all the higher level processes (i.e. navigation, autonomy, control and vehicle and missions safety algorithms) are agnostic to the computer board. 

\subsection{Sensor software drivers within MITFrontseat}

As seen from Figure \ref{fig:complete_system}, an array of MOOS drivers were developed to communicate with various hardware sensors and actuators used in the vehicle via various hardware interfaces available onboard the BeagleBoard. Each sensor driver publishes the raw sensor data to the MOOS database of the {\textit MITFrontseat} MOOS community.

\subsubsection{Depth sensor driver ({\tt iBlueRoboticsDepth})} \hfill

The depth of the vehicle was obtained by measuring the external hydrostatic pressure, and converting it to a corresponding depth value, accounting for the water temperature and density. The \textit{Morpheus} vehicle was equipped with a Blue Robotics depth sensor \cite{bluerobotics_30bar} that housed an MS5837-30BA pressure sensor \cite{MS5837_30BA}. The pressure sensor was connected to the I$^2$C bus of the BeagleBone Blue embedded computer using its JST connectors.

A MOOS driver application, {\tt iBlueRoboticsDepth}, was developed to read the external pressure and temperature measurements from the Blue Robotics pressure sensor, and to compute the corresponding vehicle depth, using a pre-configured water density. 
The calculated depth was then filtered with an outlier rejection scheme -- if a depth reading that suggests a depth rate of over 5 m s$^{-1}$ (pre-configurable) was observed, it is rejected since a 5 m s$^{-1}$ depth rate is unrealistic. Upon the outlier rejection scheme, a moving average filter with a window size of 20 samples (pre-configurable), which results in a time interval of around 2 seconds, is also applied to smoothen the data. Filtered depth value is finally published to the MOOS database.

\subsubsection{IMU sensor driver ({\tt iBBBlue} and {\tt iXsensMTi})} \hfill

The attitude (i.e. the roll, pitch and heading angles) of the vehicle was measured using an InvenSense MPU-9250 \cite{MPU_9250} micro-electromechanical system (MEMS) 9-axis inertial measurement unit (IMU), embedded in the BeagleBone Blue computer board, which is routed to the I$^2$C bus. 

An existing, thirdparty MOOS driver application, {\tt iBBBlue} \cite{projectladon} was used to read data from the IMU. The {\tt iBBBlue} application utilizes several functions given in the Robot Control Library \cite{librobotcontrol_manual} to conduct tasks such as reading IMU data from the I$^2$C bus, IMU calibration correction and fusion of acceleration, angular velocity and magnetic intensity data to compute the  roll, pitch and heading of the sensor. These raw data are then published to the MOOS database.

For applications that require more accurate attitude and heading information as compared to the InvenSense MPU-9250 \cite{MPU_9250} sensor, an external IMU sensor can be utilized, together with a MOOS driver for the sensor. An example is the Xsens MTi-3 \cite{xsens_mti3} IMU unit, which can be connected to the BeagleBone using the UART interface. A MOOS driver, {\tt iXsensMTi}, was created to read fused attitude data from the sensor, and publish them to the MOOS database.

\subsubsection{GPS and cellular modem driver ({\tt iAdafruitFona})} \hfill

An Adafruit FONA cellular breakout board \cite{adafruit_fona_3g}, which contains a SIM5320 cellular module with an integrated GPS receiver \cite{sim5320_manual} was used as the GPS and cellular modem of the vehicle. This module is connected to the vehicle's BeagleBone Blue computer using the UART interface. A MOOS driver application, {\tt iAdafruitFona}, was developed to publish raw GPS data to the \textit{MITFrontseat} MOOS database. This application also acts as a service that sends and receives short message service (SMS) text messages. Any incoming SMS messages from allowed phone numbers (pre-configured) are published to the MOOS database with the message content and sender's phone number. Any application within the \textit{MITFrontseat} community can publish a specific MOOS variable to the database, containing the message content and phone number to forward the message to an outside phone number. For example, this service can be used to send an SMS to the vehicle operator's phone number with the GPS coordinates, upon mission completion and surfacing.

While the SIM5320 module also has the internet tethering capability, which could enable remote login to the BeagleBone Blue computer via the cellular network, this was not implemented in this work.

\subsubsection{Battery and power management system driver ({\tt iBBBlue})} \hfill

A custom battery and power management system was developed and integrated to monitor the main motor current draw and main battery voltage. The current and voltage values are provided to the BeagleBone Blue via its analog-to-digital converters (ADC). The {\tt iBBBlue} reads these values are publishes them to the \textit{MITFrontseat} MOOS community, which is used by {\tt pFrontseatManager} for vehicle safety management.   

\subsubsection{Monitoring the internal pressure} \hfill

Typically, the air inside the vehicle pressure hull is partially pumped out through a vacuum port (see Section \ref{sec:nosecone}) upon vehicle assembly. Once the air is pumped out, the vacuum port is closed, leaving a low pressure zone inside the hull. If the hull is watertight, the internal pressure will be holding. A raise in the internal pressure indicates a leak in the pressure hull.

In the base vehicle, the air pressure inside the pressure hull is measured using the barometer embedded in the BeagleBone Blue computer. The barometer reading can be monitored upon vehicle assembly using functions and scripts given in the Robot Control Library \cite{librobotcontrol_manual}. 

\subsection{Navigation software - HydroMAN 2.0}

In our software architecture, the \textit{MITFrontseat} outsources the navigation task to an independent navigation engine, similar to most commercial AUVs (i.e. most AUVs rely on commercially available INS units for navigation -- an INS is a combination of an IMU and a computer running a `black-box' navigation fusion algorithm that fuses the IMU measurements with external sensors such as GPS, DVL, depth, USBL, etc. \cite{bluefin_21,ISE_explorer,Teledyne_gavia}). In this work, we utilize \textit{HydroMAN 2.0} as the navigation engine.

\textit{HydroMAN} (stands for hydrodynamic model aided navigation) is a self-learning, independent underwater navigation engine \cite{randeni_hydroman,randeni_PMS406,randeni_MOOSDAWG_hydroman_frontseat_vectors,randeni2020construction,randeni2020dynamic,randeniAUV18}. The HydroMAN 2.0 synthesizes raw measurements from sensors such as IMU, DVL, CVL, LBL/USBL, terrain-aided navigation and GPS into its self-calibrating vehicle flight dynamic model to compute the navigation solution, with the use of an array of sensor pre-processors and a layered extended Kalman filter based fusion algorithm. When accurate sensor measurements are available; for example, DVL bottom-lock and/or acoustic position updates, the HydroMAN self-calibrates the vehicle model to the local operating environment, largely compensating for the navigation drift provided by underwater currents and the flight dynamic model's own error estimate.
The calibrated vehicle model is then utilized for navigation aiding when accurate sensors are unavailable, or turned off in order to save power.

Expensive tactical and navigation grade INS units and DVLs are infeasible for low-cost micro-AUVs that are limited by the cost, such as \textit{Morpheus}. Therefore, they typically rely on inexpensive MEMS IMUs and an RPM-to-speed table for dead-reckoning based navigation, resulting in poor navigation performance. For such vehicles, \textit{HydroMAN} uses a pre-identified vehicle flight dynamic model to estimate the vehicle velocity in surge, sway and heave directions, by using a combination of input variables such as the propeller speed, control surface angles and the rates of changes of roll, pitch and heading angles. As shown in \cite{randeniAUV18} and \cite{randeni2020dynamic}, the vehicle dynamic model is capable to improving the navigation accuracy by orders of magnitude as compared to traditional RPM-to-speed curve based dead-reckoning. In the meantime, \textit{HydroMAN} also provides the possibility to extend the base vehicle with additional navigation sensors if one anticipates to, without requiring changes to the navigation software stack.

HydroMAN is comprised of a number of MOOS applications as described in this companion paper \cite{randeni_hydroman}. Following subsections briefly summarize key functionalities of these sub-component:

\subsubsection{The interface to the HydroMAN MOOS community ({\tt iHydroMAN\_Gateway})} \label{sec:hydroman_gateway} \hfill

The \textit{HydroMAN} version 2.0 is an independent navigation engine that interfaces with client systems (i.e. \textit{MITFrontseat} in this case) using a TCP network connection. Therefore, the front-end of \textit{HydroMAN 2.0} is similar to internal fusion engines of tactical and navigation grade INSs -- the client system can send external raw sensor data to \textit{HydroMAN}; and \textit{HydroMAN} will provide the fused, model-aided navigation solution. Hence, the \textit{HydroMAN} is independent of the client's software architecture.

{\tt iHydroMAN\_Gateway} application runs within the HydroMAN MOOS community, serving as the gateway to \textit{HydroMAN}. All incoming messages to HydroMAN (e.g. raw sensor data) and outgoing messages from HydroMAN (e.g. final navigation solution) are handed over by {\tt iHydroMAN\_Gateway}. It runs a TCP server, which allows the client system's HydroMAN driver (in this case, {\tt iHydroMAN\_MITFrontseat} MOOS application that runs within the \textit{MITFrontseat} MOOS community) to connect as a TCP client, and exchange messages using a google protocol buffer based standardized message definition. This standardized message definition and server-client architecture ensures the independence of HydroMAN; i.e., the client system's HydroMAN driver does not necessarily need to be a MOOS application. 

\subsubsection{Vehicle flight dynamic model ({\tt pHM\_BasicModel})} \hfill

As virtual navigation aiding sensor, an embedded AUV flight dynamic model, based on the principle of conservation of energy \cite{randeni2020dynamic} is used in HydroMAN to estimate the linear velocities of the AUV (i.e. $u$, $v$ and $w$). While the actual structure of the vehicle dynamic model varies from vehicle to vehicle, Equations \ref{eq-energy_u}-\ref{eq-energy_w} demonstrate how the propeller speed ($RPM_{(t)}$), measured vehicle angular velocities ($p_{(t)}$, $q_{(t)}$ and $r_{(t)}$), and the linear velocities estimated at the previous timestamp ($u_{(t-1)}$, $v_{(t-1)}$ and $w_{(t-1)}$) are used to derive the vehicle velocities at the current timestamp (i.e. $u_{(t)}$, $v_{(t)}$ and $w_{(t)}$) for an example vehicle:

\begin{equation}
\begin{split}
u_{(t)}= \alpha_{1}RPM_{(t)} + \alpha_{2}RPM_{(t)}^2 +\alpha_{3} q_{(t)} p^2  + \\
\alpha_{4}r_{(t)} v_{(t-1)}^2 + \alpha_{5}q_{(t)} w_{(t-1)}^2 + \alpha_{6} p_{(t)}^2 + \alpha_{7} q_{(t)}^2 + \\
\alpha_{8} r_{(t)}^2 + \alpha_{9} p_{(t)} r_{(t)}^2 + \alpha_{10} z_{(t)}
\label{eq-energy_u}
\end{split}
\end{equation}

\begin{equation}
\begin{split}
v_{(t)}=  \beta_{1} q_{(t)} p_{(t)}^2 + \beta_{2} p_{(t)}^2 + \beta_{3} r_{(t)} u_{(t-1)}^2 + \\ \beta_{4} q_{(t)} u_{(t-1)}^2 +  \beta_{5} q_{(t)}^2 + \beta_{6} r_{(t)}^2 + \beta_{7} p_{(t)} r_{(t)}^2 + \beta_{8} z_{(t)}
\label{eq-energy_v}
\end{split}
\end{equation}

\begin{equation}
\begin{split}
w_{(t)}=  \gamma_{1} r_{(t)}u_{(t-1)}^2 + \gamma_{2} q_{(t)} u_{(t-1)}^2 + \gamma_{3} q_{(t)} p_{(t)}^2 + \\
\gamma_{4} p_{(t)}^2 + \gamma_{5} q_{(t)}^2 + \gamma_{6} r_{(t)}^2 + \gamma_{7} p_{(t)} r_{(t)}^2   + \gamma_{8} z_{(t)}
\label{eq-energy_w}
\end{split}
\end{equation}

\noindent where $\alpha_n$, $\beta_n$, and $\gamma_n$ are AUV-dependent flight dynamic model parameters that were estimated using a real-time recursive least squares system identification algorithm.
The derivation of the dynamic model, model optimization and parameter estimation procedures are beyond the discussion of this paper, and further details are given in \cite{randeni2020dynamic}.

The velocity and position estimated by the flight dynamic model are relative to the water column (i.e. $\nu ^{{model}}_{\mathrm{(auv|water)}}$ and $x ^{{model}}_{\mathrm{(auv|water)}}$) since the model excludes water currents.
Therefore, the error sources of the model-based velocity and position estimates include the drift due to water currents and the uncertainty of the model \cite{randeniAUV18,randeni2020dynamic}, which are counteracted by the self-adaptation of the flight dynamic model

\subsubsection{Self-calibration of the flight dynamic model to the operating environment ({\tt pHM\_ModelCalibrator})} \hfill

The uncertainty of the dynamic model, and water current velocity ($\nu_{\mathrm{(water|earth)}}$) are estimated on-the-fly within {\tt pHM\_ModelCalibrator} when accurate sensor measurements such as DVL bottom-lock and/or acoustic navigation updates are available. These estimates are used to convert the model-based velocity from water reference to earth reference, as detailed in Equation \ref{eq-auv_rel_water}:

\begin{equation}
\begin{split}
\nu ^{{adaptM}}_{\mathrm{(auv|earth)}} = \nu ^{{model}}_{\mathrm{(auv|water)}} + \nu ^{{adaptM}}_{\mathrm{(water|earth)}} + \sigma ^{adaptM}_{\mathrm{\nu ^{model}}}
\label{eq-auv_rel_water}
\end{split}
\end{equation} 

Two self-calibration strategies are available within {\tt pHM\_ModelCalibrator}: (a) using acoustic position updates and (b) using the bias error of the model-based velocity (i.e. bias error estimated by the error-state extended Kalman filter (EKF) of the fusion algorithm). Further details on these algorithms are given in \cite{randeni_hydroman}.

\subsubsection{DVL pre-processor ({\tt pHM\_DVLProcessor})} \hfill

This MOOS application processes raw sensor measurements from velocity aiding sensors such as DVL and CVL. By considering the configured orientation of the sensor, the velocity measurements are transformed to the \textit{HydroMAN} standard axis convention. An orientation mismatch detection mechanism is implemented to warn the operator and/or execute vehicle safety protocols if the configured sensor orientation is detected to be not accurate. 

In under-ice operations, if the sensor is in an upward-facing configuration, measuring the velocity of the AUV relative to surface ice, {\tt pHM\_DVLProcessor} is capable of counteracting the velocity for potential drifts in the surface ice (i.e. surface ice in the Arctic is translated and rotated by wind and current forcing \cite{mcphee1980analysis}, and this ice drift velocity can be up to around 1 m s\textsuperscript{-1}, which can cause considerable navigation drift \cite{kaminski201012}). {\tt pHM\_DVLProcessor} can be aided with ice drift velocity information obtained from actual measurements (e.g. measured by a GPS unit located on the surface, and transmitted down to the vehicle via an acoustic link) or from modeling approaches \cite{mo2020simulation}. 

\subsubsection{LBL pre-processor ({\tt pHM\_LBLProcessor})} \hfill

Navigation aiding information provided in the form of position updates; for examples, acoustic position updates from LBL/USBL/SBL systems, terrain-aided navigation updates, etc. are pre-processed by the {\tt pHM\_LBLProcessor} application.

Some types of acoustic position updates (e.g. two-way-travel-time systems) can typically be outdated by more than 20 s when the position update is received by the AUV (i.e. $t-t_{N}$, where $t$ is the current timestamp). A 20-second time-lag could develop a position error of up to around 32 m (assuming a speed of 1.6 m s\textsuperscript{-1}); hence, are typically rejected by most commercial INS sensor fusion algorithms. {\tt pHM\_LBLProcessor} contains an algorithm to extrapolate such position updates to the current timestamp using the self-adapting vehicle flight dynamic model as given in Equation \ref{eq-lbl_at_t}:

\begin{equation}
\begin{split}
x ^{{lbl}}_{\mathrm{(t)}} = x ^{{lbl}}_{\mathrm{(t_{N})}} + \left( x ^{{adaptM}}_{\mathrm{(t)}} - x ^{{adaptM}}_{\mathrm{(t_{N})}} \right)
\label{eq-lbl_at_t}
\end{split}
\end{equation}

\noindent where $x ^{{adaptM}}_{(T)}$ is the vehicle position from the self-adapting flight dynamic model at timestamp $T$. The current timestamp is given by $t$ and the LBL timestamp is given by $t_{N}$.

This pre-processor allows \textit{HydroMAN} to effectively utilize navigation updates that are time-lagged by large time periods (i.e. more than 5-minutes). 

\subsubsection{Sensor fusion engine ({\tt pHM\_SensorFusion})} \hfill

The sensor fusion application consists of two EKFs: (a) the error-state EKF that estimates the bias errors of the sensors, and (b) main-state EKF that fuses the bias error corrected measurements to obtain the final navigation solution.

The error-state EKF computes a running estimate of the bias errors of velocity sensors (e.g. DVL, CVL, etc.) and flight dynamic model in a layered pattern, in the hierarchy of the accuracy of the sensor.
That is, the outlier removed acoustic position updates are first used to compute the bias error estimate of the DVL sensor, which is used to correct the DVL measurements. The bias error corrected DVL measurements are then used to compute the bias error of dynamic model, and other velocity aiding sensors, in a hierarchical order. Since the variation of bias error is generally a slowly changing function, this method allows \textit{HydroMAN} to maintain a good navigation accuracy by using bias corrected dynamic model, even in events where the DVL drops out or turned off for a long period of time.

The main-state EKF included six states --- the three dimensional velocity and position vectors. These states were estimated by fusing the bias corrected DVL, flight dynamic model and other velocity measurements together with the depth and position based navigation updates.
More information regarding this layers sensor fusion approach is given in \cite{randeni_hydroman}.

\subsubsection{Navigation manager ({\tt pHM\_Manager})} \hfill

The \textit{HydroMAN} system consists of a number of navigational safety management systems; e.g. EKF re-initialization when large navigation drifts are detected, executing safety protocols in situations where the filter is diverging due to faulty sensors, etc. The {\tt pHM\_Manager} application manages these features and publishes the final navigation solution.

\subsection{Autonomy software}

Unlike unmanned sea-surface, ground, and aerial vehicles, AUVs cannot be remotely controlled due to the low bandwidth in acoustic communications; they must make decisions autonomously. Remote control, or teleoperation, in land, air, or surface vehicles may be viewed as a means to allow conservative, risk-averse operation with respect to the degree of autonomy afforded to the vehicle. In underwater vehicles, similar conservative tendencies are realized by scripting the vehicle missions to be as predictable as possible. Missions typical of early-model UUVs were composed of a preplanned set of waypoints accompanied by depth and perhaps speed parameters. The onboard sensors merely collected data that were analyzed after the vehicle was recovered from the water. However, improved sensor processing methods, embedded computing power, underwater navigation performance and adaptive and collaborative autonomy technology has enabled advanced autonomy for AUVs \cite{benjamin2010nested}.

The base vehicle software stack that we developed carries several autonomy capabilities of several fidelity levels: (1) primitive missions with scripted decision outputs; (2) autonomous decision making with MOOS-IvP behavioural helm that runs on the \textit{MITFrontseat} MOOS community; and (3) payload autonomy where the \textit{MITFrontseat} ingests decision commands from thirdparty payload-based autonomy systems. 

\subsubsection{Higher level autonomy management \\ ({\tt pFrontseatManager})} \hfill \label{sec:pFSM}

The autonomy helm of the vehicle, regardless of the fidelity level, sits beneath and bound by a safety envelope set by this mission management application. In addition to enforcing safety rules, this application is also responsible for executing and switching autonomy behaviors with the use of a state machine. 

The frontseat mission manager enforces vehicle-dependant and cruise-dependant safety rules, set by the operator during pre-launch mission configuration. The vehicle-dependent rules ensure the integrity of structural and electrical components and water-tightness of the vehicle by administering variables such as the maximum vehicle diving depth, minimum operating battery voltage, maximum operating motor current, maximum internal pressure. When a specific rule is violated, the vehicle mode will be autonomously switched to an orchestrated safe mode, depending on the violated rule. The cruise, or mission dependent rules include: (a) mission start time -- the main motor start time could be delayed by a pre-configured time period since the mission launch; (b) mission end time -- the mission manager leases the vehicle's control authority to the autonomy helm only for a pre-configured temporary time period; beyond which, the mission ending mode is executed; (c) maximum cruise depth -- if the maximum safe operating depth for the cruise region is below the maximum diving depth of the vehicle. 

When MOOS-IvP helm is run within the \textit{MITFrontseat} MOOS community, the frontseat mission manager functions as a mission commander that carries out on-board command and control of mission behaviors. That is, with the use of a state machine, this application controls which IvP helm behaviors are spawned at a given time \cite{benjamin2010nested} and how they are switched between. The switching of IvP behaviors is either conducted completely autonomously, or manually triggered via communication methods detailed in Section \ref{sec:comms}. 

\subsubsection{Passive helm ({\tt pHelmPassive})} \hfill

The passive helm allows scripting of a timetable of pre-defined helm decisions (i.e. desired speed, desired heading, desired depth and vehicle mode command); each against a corresponding execution time (i.e. mission legs). During the mission, {\tt pHelmPassive} reads the pre-configured helm decisions from the configuration block, and posts to the MOOS database. Therefore, this primitive helm can be run without a vehicle navigation solution, making it a useful tool for preliminary testing of the vehicle.

Another key use case of the passive helm is for tuning of the vehicle's low-level control system. Most AUVs still use proportional-integral-derivative control systems for their low-level control; and fine-tuning them, which is typically done trial-and-error, is rather dull process. During this process, the autonomy system is required to first command a constant heading, speed and depth; followed by a step-change command in the either heading, speed or depth, depending on which degree-of-freedom is being tuned. The passive helm is an ideal tool for such simple, pre-dictated missions. In addition, passive helm also allows the users to configure PID gain changes during legs, which is ingested by the control engine as runtime PID gain updates. This functionality allows the operators to test multiple PID gain settings during a single mission, expediting the time consuming tuning process by orders of magnitude. Listing \ref{listing:pHelmPassive} shows a sample mission configuration block of {\tt pHelmPassive}, where the P-gain of the vehicle's heading controller is updated during the third leg.

\begin{lstlisting}[caption=A sample mission configuration block of {\tt pHelmPassive}. This primitive mission updates the proportional gain of the vehicle's heading PID controller during the third leg., captionpos=b, label={listing:pHelmPassive}]
ADD_LEG: start_time=120, heading=180, speed=1.5, depth=1.5
ADD_LEG: start_time=240, heading=250, speed=1.5, depth=2.0
ADD_LEG: start_time=410, heading=250, speed=1.5, depth=2.0, heading_kp=0.8 
ADD_LEG: start_time=420, heading=180, speed=1.5, depth=2.0 
ADD_LEG: start_time=600, heading=250, speed=1.5, depth=1.5
\end{lstlisting}

\subsubsection{MOOS-IvP autonomy helm ({\tt pHelmIvP})} \hfill

The MOOS-IvP helm runs as a single MOOS application and uses a behavior-based architecture for implementing autonomy. Behaviors are distinct software modules that can be described as self-contained mini-expert systems dedicated to a particular aspect of overall vehicle autonomy. The helm implementation and each behavior implementation expose an interface for configuration by the user for a particular set of missions. This configuration often contains particulars such as a certain set of waypoints, search area, and vehicle speed. It also contains a specification of mission modes that determine which behaviors are active under which situations and how states are transitioned. When multiple behaviors are active and competing for influence of the vehicle, the IvP solver is used to reconcile the behaviors. More information regarding MOOS-IvP can be found from  \cite{benjamin2010nested} and \cite{moosivp_web}.

The \textit{Morpheus} base vehicle software stack allows the users to run MOOS-IvP autonomy from within the \textit{MITFrontseat} MOOS community. In this architecture, required behaviors are loaded to the mission configuration block, and the {\tt pFrontseatManager} application acts as the mission commander in-charge of spawning and switching between behaviors. 

\subsubsection{Payload autonomy} \hfill \label{sec:payload_autonomy}

The main idea in the payload autonomy paradigm, or the backseat driver is the separation between vehicle control and vehicle autonomy. The vehicle control system runs on a platform’s main vehicle computer, and the autonomy system runs on a separate payload computer. This separation is also referred to as the mission controller – vehicle controller interface. A primary benefit is the decoupling of the platform autonomy system from the actual vehicle hardware \cite{benjamin2010nested,balasuriya2010behavior,eickstedt2009backseat,viquez2016design,naglak2018backseat,hwang2020enhancement,underwood2017design}.

The payload autonomy capability is built-in to the \textit{Morpheus} base vehicle software stack through the development of a standard payload interface. This interface allows to forward any information posted in the \textit{MITFrontseat} MOOS community to a payload computer via a TCP connection, using a standard protobuf message scheme; and the payload autonomy system is able to send autonomy commands (e.g. desired heading, desired speed and desired depth commands) back to the \textit{MITFrontseat} via the same interface. More information regarding the interface is given in Section \ref{sec:payload_interface}. Hardware-wise, the payload autonomy system could run either on the same main vehicle computer or on a separate autonomy computer. In the case of \textit{Morpheus}, the payload autonomy system was also run on the same main vehicle computer (i.e on the BeagleBone Blue) in order to conserve space inside the vehicle.

In the payload autonomy mode, as discussed in Section \ref{sec:pFSM}, the {\tt pFrontseatManager} leases the command of the vehicle to the payload autonomy system for a pre-configured time period. However, the payload autonomy system is still bound by the safety envelope set by the frontseat manager. If one anticipates to take unconditional control of the vehicle, this can be done by turning off safety parameters within {\tt pFrontseatManager}.

\subsection{Communication software} \label{sec:comms}

Communication of low-cost, micro AUVs such as \textit{Morpheus} can be generally classified into three categories: (1) short-range surface communication; (2) long-range surface communication; and (3) underwater acoustic communication. 

Short-range surface communication is generally via a wifi network connection with the topside network. \textit{Morpheus} vehicle achieves this using the BeagleBone Blue computer's embedded wifi modem. The computer's network settings are configured such that it connects to a specific wifi network whenever the vehicle is in range. This network is typically used to access the vehicle computer in order to conduct operations such as system testing, launching missions, debugging, data transfer, etc.

In \textit{Morpheus} vehicle, long-range surface communication is achieved via the cellular network; with the use of SMS messages with a dedicated topside cellular phone. Very basic command and control, and vehicle status monitoring can be achieved with this service (e.g. this service is typically configured to send an SMS to the topside phone with the GPS coordinates, upon mission completion and surfacing). We expect to expand the long-range surface communication capability by establishing a remote connection between the topside computer and vehicle with the use of cellular internet tethering for more advanced surface command and control, and telemetry monitoring; though the use of Goby-Acomms library \cite{schneidergobyAcomms,schneider2012goby} for marshalling and dynamic priority queuing; and Goby liaison as the command and control GUI \cite{goby3_manual}.

At the time of writing, the \textit{Morpheus} vehicle is not equipped with an acoustic modem that enables transmission of datagrams while underwater. However, the piUSBL system in the \textit{Perseus} payload allows very basic underwater command and control by switching the transmitter between various broadcast linear frequency-modulated chirps; each corresponding to different pre-defined vehicle autonomy behaviors \cite{rypkema2021synchronous}. In the future, we plan to expand the electronics stack of the vehicle with a small acoustic modem (e.g. \cite{freitag2005whoi,won2012design,renner2020ahoi}) for more advanced underwater command and control, and telemetry; through the use of Goby-Acomms library, which will also be used for long-range surface communication.  

\begin{figure*}[h]
\centering
\includegraphics[trim=0 0 0 0, clip,  width=1.0 \textwidth]{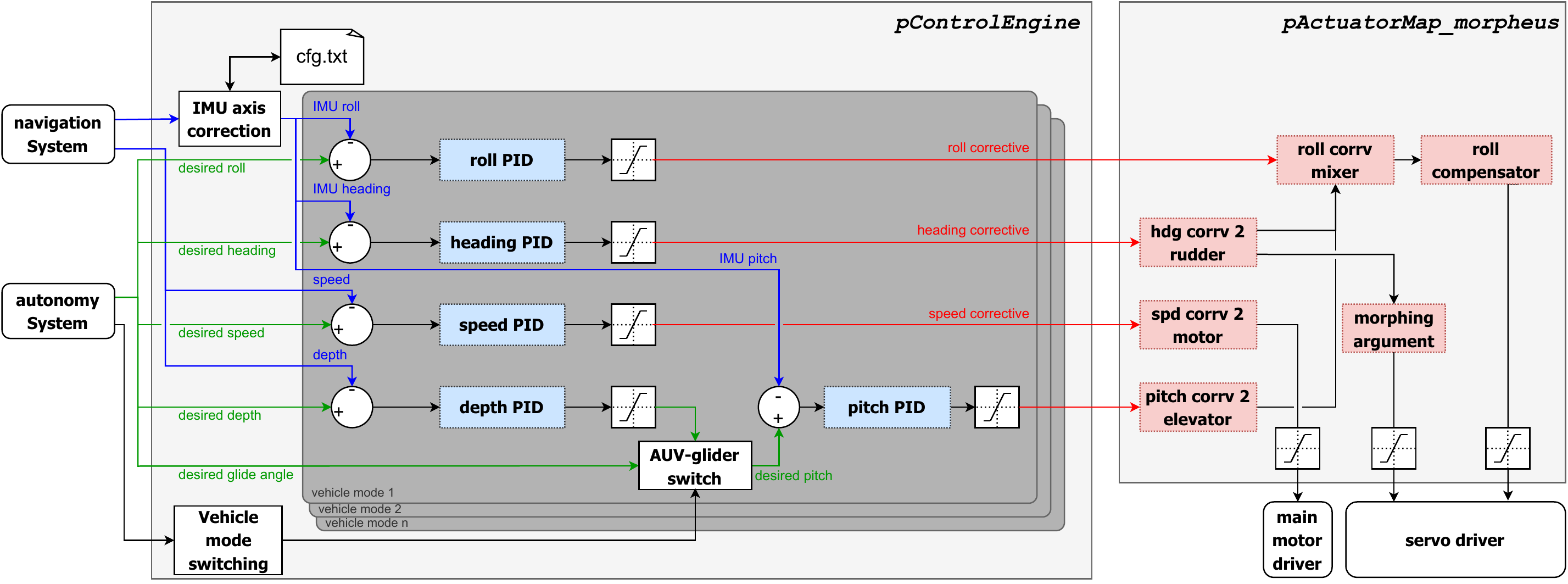}
\caption{The low-level control system of \textit{MITFrontseat} has been semi-generalized by sub-dividing it to three components: (1) a platform-independent control engine ({\tt pControlEngine}) that produces control correctives in roll, heading, speed, and pitch; (2) a platform-dependant actuator mapper application (e.g. {\tt pActuatorMap\_morpheus}) that maps the control engine outputs to the actuator configuration of a specific vehicle; and (3) actuator drivers that produces low-level signals such as PWM, GPIO and CAN bus messages that drive the actuators.}
\label{fig:low-level_control}
\end{figure*}

\subsection{Low-level control software}

The low-level control software is responsible for executing the decisions commanded by the autonomy system; such as, desired heading, desired speed, desired depth and desired glide angle (i.e. in the case for gliding vehicles) commands. The vehicle control system executes such commands by controlling the vehicle-specific actuators such as the propulsion thrusters, control surfaces, buoyancy engines and weight shifting mechanisms, etc. Hence, the low-level control system of an AUV is generally platform-dependant. In \textit{MITFrontseat}, we have generalized the control system by sub-dividing it to three components as shown in Figure \ref{fig:low-level_control}: (1) a platform-independent control engine that produces control correctives in roll, heading, speed, and pitch; (2) a platform-dependant actuator mapper application that maps the control engine outputs to the actuator configuration of a specific vehicle; and (3) actuator drivers that produces low-level signals such as PWM, GPIO and CAN bus messages that drive the actuators.

\subsubsection{Control engine ({\tt pControlEngine})} \label{sec:controlengine} \hfill

The low-level control engine of \textit{MITFrontseat} consists of a set of single loop (i.e for heading, speed and roll sub-systems) and multi-loop (i.e. for the depth sub-system) PID control blocks that produce control corrective outputs. Control corrective outputs are essentially in the same order as actuator commands; e.g. control surface angle commands. In order to ensure the platform-independence of the control engine, PID outputs are published as control correctives; and the mapping of control correctives to actuators of a specific vehicle is conducted in a separate MOOS application.

For some vehicle hardware designs, there could be an offset between the IMU mount axis and vehicle axis. In \textit{MITFrontseat}, this offset correction is carried out in the control engine. The axis offset correction for heading is done by pointing the vehicle's nose towards north, and poking a given MOOS variable. Similarly, roll and pitch offsets are corrected by keeping the vehicle at zero roll and pitch, and poking two separate MOOS variables. These offsets are written to a configuration text file, which is read on start-up to correct the IMU offset.

The control engine contains three independent single-loop PID control blocks for heading, speed and roll sub-systems. They ingest the difference between the desired and actual values (i.e. for example, the heading error), and compute a PID corrective that would attempt to minimize the error, with the use of configured PID gains. 

For under-actuated vehicles such as flying type AUVs, the depth DOF cannot be directly controlled; and is rather controlled by varying the pitch angle of the vehicle. Thus, a two-loop PID controller is implemented for the depth sub-system. As seen from Figure \ref{fig:low-level_control}, the depth PID control block produces a depth control corrective, which becomes the desired pitch input for the pitch PID control block. The latter then computes the pitch control corrective, which is sent to the actuators that control the pitch DOF of the vehicle (e.g. elevators). However, for gliding vehicles, the optimized desired glide angle (i.e. desired pitch value) is provided by the autonomy system. For such vehicles, the control engine by-passes the depth PID block; and the pitch PID block uses the desired glide angle as the desired pitch value.

The \textit{MITFrontseat} is compatible for vehicle with multiple modes; for example, for hybrid gliders that has both propelled and gliding modes; and for amphibious vehicles that are capable of operating in-water as well as ashore. For such multi-mode vehicles, multiple control settings are typically required; for instance, in the case of hybrid vehicles, one PID setting for the propelled mode (i.e. regulating the propeller and control surface angles to control the speed, pitch and heading), and another PID setting for the gliding mode (i.e. regulating the buoyancy engine and battery-pack position to control the same variables). As shown in Figure \ref{fig:low-level_control}, the control engine handles this by dynamically creating an `N' number PID controller sets during start up, each set corresponding to a vehicle mode. When the autonomy system switches to a specific vehicle mode, the control engine also switches itself to the corresponding PID controller and gain setting.

\subsubsection{Mapping control correctives to vehicle actuators ({\tt pActuatorMap\_morpheus})} \label{sec:pactuatormap} \hfill

In this framework, as shown in Figure \ref{fig:low-level_control}, the platform-independent control correctives produced by the control engine are converted to actuator commands of a given vehicle by a platform-dependent MOOS application; for instance, in the \textit{Morpheus} AUV, this is carried out by {\tt pActuatorMap\_morpheus}. 
As outlined in Section \ref{sec:tailcone}, the \textit{Morpheus} class vehicles are equipped with four independently actuated stern control surfaces (i.e. upper rudder, lower rudder, port elevator and starboard elevator), two vertical forward morphing fins and a propeller at the stern end. 

\begin{figure}[h]
\centering
\includegraphics[trim=0 0 0 0, clip,  width=0.5 \textwidth]{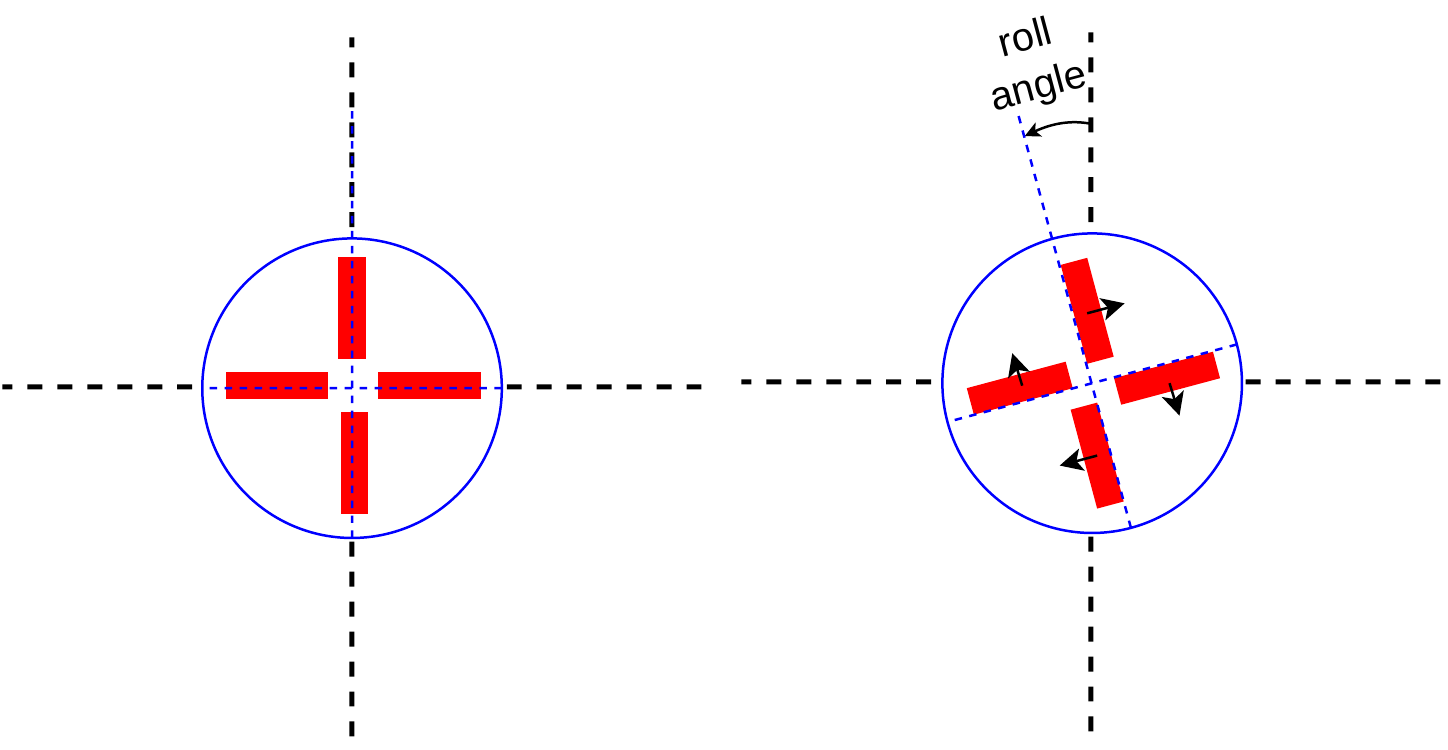}
\caption{Roll correction -- Left: when the AUV is at zero roll, the control surfaces are at their neutral positions. Right: when the vehicle is rolled, the zero position of all four control surfaces are offset by a small angle, creating a righting moment to correct the vehicle roll angle back to zero.}
\label{fig:roll_correction}
\end{figure}

The heading and pitch correctives are first mapped out to corresponding stern rudder and elevator angles. If the vehicle is at zero roll angle (or if the roll control subsystem is deactivated), these commands will be the final rudder and elevator commands. However, in situations where the vehicle is rolled, as shown in Figure \ref{fig:roll_correction}, the zero position of all four stern control surfaces will be offset by a small angle (the maximum deflection limit is typically configured as 5 degrees), attempting to create a righting moment to zero out the roll angle.

\begin{figure}[h]
\centering
\includegraphics[trim=0 0 0 0, clip,  width=0.5 \textwidth]{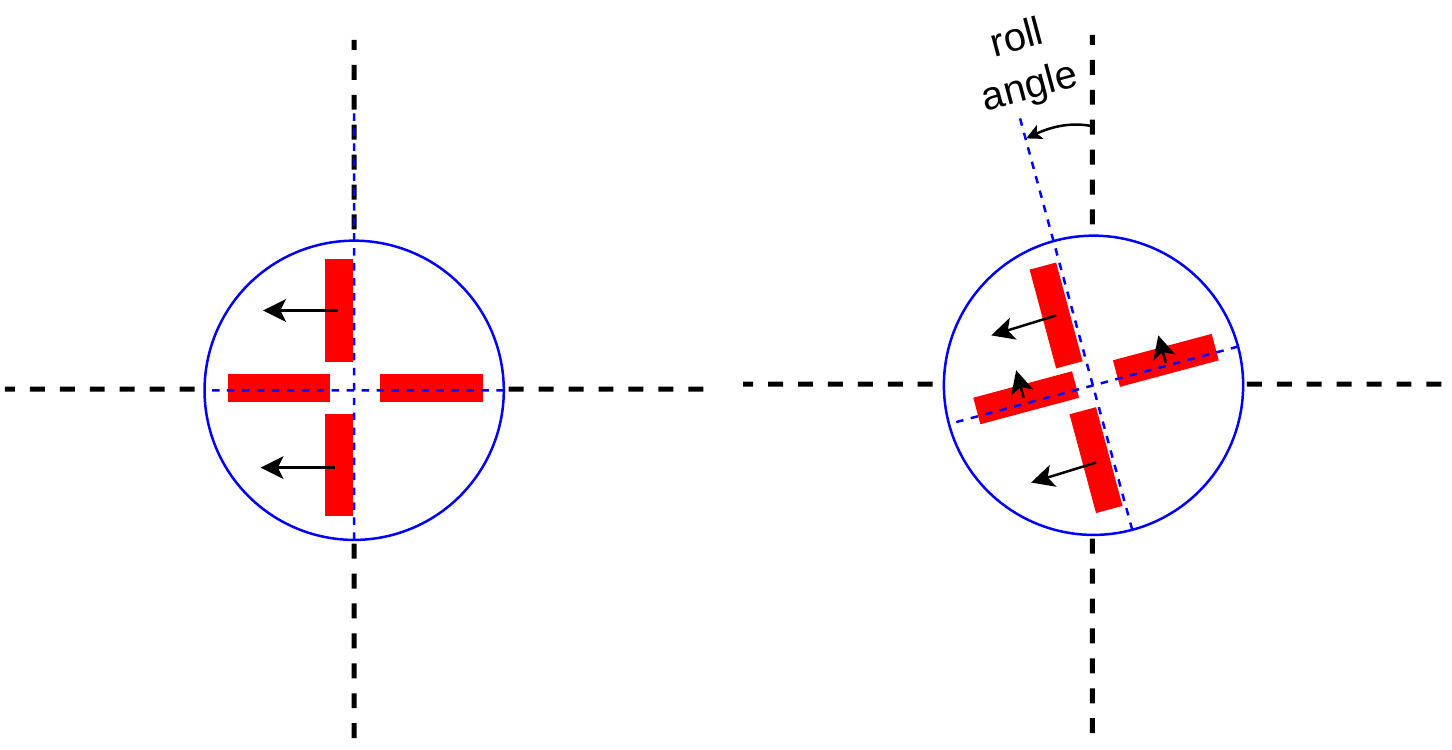}
\caption{Roll compensation -- Left: when the AUV is at zero roll, heading correction is simply mapped out to a rudder deflection. Right: when the AUV is rolled, however, a simple rudder deflection will not only create a heading change, but also will create an unintended pitch change. Roll compensating system will attempt to mitigate this by accordingly deflecting the elevators to cancel out the pitching moment created by the rudders, and vice versa.}
\label{fig:roll_compensation}
\end{figure}

As shown in Figure \ref{fig:roll_compensation}, when the vehicle is at non-zero roll angles, a rudder deflection will not only create a heading change, but will also create an unintended pitching moment, and vice versa. Thus, the vehicle will have unintended depth fluctuations during turns, and heading fluctuations during depth changes. The roll compensation system within {\tt pActuatorMap\_morpheus} attempts to mitigate this by accordingly deflecting the opposing control surfaces to cancel out the unintended moment as given in Equations \ref{eq-roll_comp_upRud} - \ref{eq-roll_comp_stbdElv}; for instance, deflecting the elevators to cancel out the unintended pitching moment created by the rudders. 

\begin{equation}
\begin{split}
uppr\_rudd = \psi ^{corr} \cos{\phi} - \theta ^{corr} \sin{\phi} + \phi^{corr}
\label{eq-roll_comp_upRud}
\end{split}
\end{equation}

\begin{equation}
\begin{split}
lowr\_rudd = \psi ^{corr} \cos{\phi} - \theta ^{corr} \sin{\phi} - \phi^{corr}
\label{eq-roll_comp_loRud}
\end{split}
\end{equation}

\begin{equation}
\begin{split}
port\_elev = \psi ^{corr} \sin{\phi} + \theta^{corr} \cos{\phi} - \phi^{corr}
\label{eq-roll_comp_portElv}
\end{split}
\end{equation}

\begin{equation}
\begin{split}
stbd\_elev = \psi ^{corr} \sin{\phi} + \theta^{corr} \cos{\phi} + \phi^{corr}
\label{eq-roll_comp_stbdElv}
\end{split}
\end{equation}

\noindent where, $\phi ^{corr}$, $\theta ^{corr}$ and $\psi ^{corr}$ are roll, pitch and heading correctives, and $\phi$, $\theta$ and $\psi$ are roll, pitch and heading angles of the vehicle, respectively. Note that equations \ref{eq-roll_comp_upRud} - \ref{eq-roll_comp_stbdElv} assume that all four stern control surfaces are equal in size and shape.

In {\tt pActuatorMap\_morpheus}, the forward located morphing fins are controlled according to the magnitude of the heading error. The fins were deployed if the heading error (i.e. the difference between the desired and current vehicle heading) is larger than 30$^{\circ}$. Once deployed, the fins were actively controlled with an equal but opposite angle to the rudder deflection. When the heading error reduced to less than 5$^{\circ}$, the morphing fins were retracted.

All final control surface angles are finally published to the MOOSDB as angle as well as normalized commands, which are to be read by the actuator drivers. The speed correctives are also mapped out and published as percentage thrust and normalized thrust commands.   

\subsection{Actuator software drivers}

A set of MOOS drivers were developed to communicate with various hardware actuators of the vehicle via hardware interfaces available onboard the BeagleBoard computer. Each driver reads corresponding commands from the MOOSDB, and drives the hardware by providing relevant GPIO, PWM and I$^2$C commands.  

\subsubsection{Main motor driver ({\tt iProp})} \hfill

The main motor MOOS driver handles main motor and its related circuitry. Main motor propeller is a hazardous sub-system; hence is protected by an electrical gate that needs to be triggered in order to switch the propeller on. During the mission envelope (which is dictated by {\tt pFrontseatManager}), The main motor driver triggers the gate by sending a GPIO signal. Subsequently, the relevant PMW signal is sent to the motor, according to the percentage thrust commanded by {\tt pActuatorMap\_morpheus}.  

\subsubsection{Servo motor driver ({\tt iServo})} \hfill

The servo MOOS driver is responsible for driving the servo motors to the positions commanded by {\tt pActuatorMap\_morpheus}. Servo driver achieves this by providing PWM signals (i.e. via the BeagleBoard's PWM channels) that incrementally changes the servo position until it arrives to the commanded position. 

\subsubsection{LED strobe driver ({\tt iLED})} \hfill

The LED MOOS driver handles the circuitry related to the vehicle's mast LED strobe. In this framework, the {\tt pFrontseatManager} posts various different LED pattern commands, each corresponding to the current mode of the vehicle. For instance, four different LED blinking patterns were configured to indicate: (1) a mission has been launched and waiting till the actuator-engage-time, (2) the mission clock is within 10-seconds to the actuator-engage-time, (3) mission is currently being executed and actuators are engaged, and (4) mission has ended and actuators are secured. The LED driver reads these LED commands and sends corresponding GPIO signals to the LED driving circuitry.

\subsection{Software extension for additional payloads and payload autonomy systems ({\tt iMITFrontseat\_Gateway})} \label{sec:payload_interface}

The base AUVs are typically extended with additional payloads according to its application \cite{hagen2002hugin,camilli2004integrating}. To ensure this extendability, the hardware as well as the software of the base vehicle should include boilerplate hooks to interface with additional payload sensors, actuators and processes \cite{hagen2002hugin}. %Via this interface, the base-vehicle typically publishes information such as the navigation solution.

%In some situations, this additional payload is an autonomy system, commonly known as a ``\textit{backseat driver}'' or a payload autonomy system, where the user is able to take control of the vehicle through their own autonomy framework \cite{benjamin2010nested,balasuriya2010behavior,eickstedt2009backseat,viquez2016design,naglak2018backseat,hwang2020enhancement}. In such situations, the base vehicle's embedded software (i.e. \textit{MITFrontseat}) forwards vehicle navigation information to the payload computer. The payload autonomy system, in return provides higher level autonomy decisions such as desired heading, desired speed and desired depth commands for the vehicle to follow. The \textit{MITFrontseat} then absorbs these commands and executes them thorough the low-level control pipeline. However, these backseat commands are still filtered by a higher level operational envelope laid out by the {\tt pFrontseatManager}; for example, safety aborts triggered by the maximum vehicle depth, maximum mission time duration, critical battery level, etc. 

The {\tt iMITFrontseat\_Gateway} is a such boilerplate hook that allows payloads (i.e. including payload autonomy systems) to connect to \textit{MITFrontseat} and exchange information. Similar to the {\tt iHydroMAN\_Gateway} discussed in Section \ref{sec:hydroman_gateway}, this application creates a TCP server, which allows payload systems to connect as a TCP client, and exchange MOOS messages wrapped around a google protocol buffer based standardized message definition. This interface allows payloads to read and publish any MOOS variable to the \textit{MITFrontseat} MOOS community. The standardized message definition and TCP server-client architecture ensures the independence of payload systems; i.e., the payload system does not necessarily need to be a MOOS based system. Multiple payload systems can be connected to \textit{MITFrontseat} at a given instance by spawning multiple instances of {\tt iMITFrontseat\_Gateway} application.

%\subsection{Mission configuration}

% ==============================================
\section{Morphing fin payload design}
% ==============================================

\begin{figure}[h]
\centering
\includegraphics[trim=0 70 180 10, clip,  width=0.5
\textwidth]{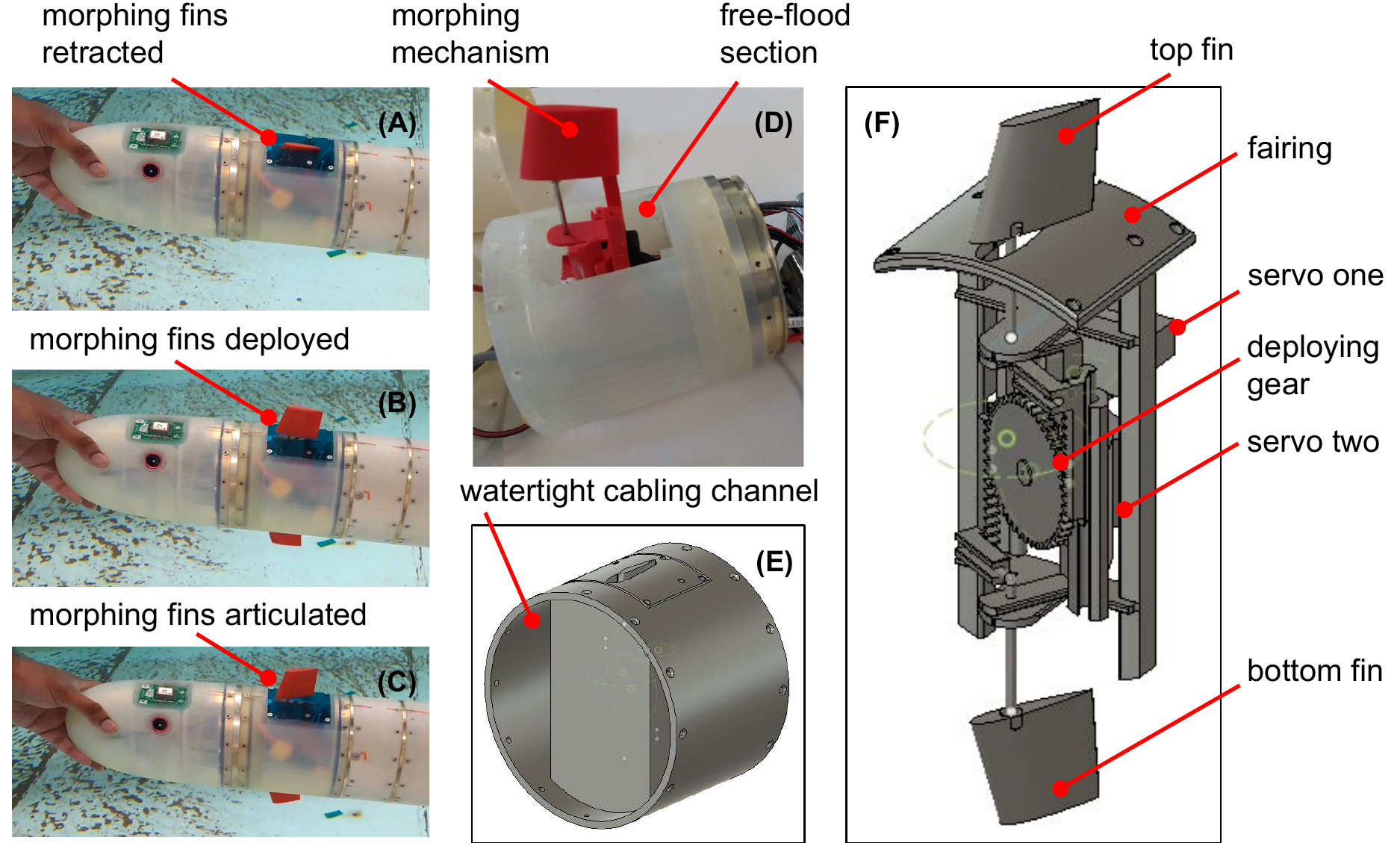}
\caption{The morphing fin payload module was placed immediately after the nose-cone of the AUV in order to obtain the maximum stability-index variation. The morphing fins can be (A) retracted, (B) deployed and (C) articulated up to a 20 degree angle of attack. (D) The morphing mechanism was housed inside a free-flood chamber within the module. (E) Two watertight channels were located on either sides of the chamber to run electrical cables across the morphing fin module. (F) The two morphing fins were driven in and out of the hull through fin cutouts by a servo-driven push rods mechanism, which was placed on a carriage that can be rotated using another servo, providing fin articulation.}
\label{fig:morphing}
\end{figure}

The stability and maneuverability indices of a torpedo-shaped vehicle can be dynamically altered using different modes of retractable fin implementations \cite{randeni2022bioinspired}. In this work, we implemented forward located morphing fins, where the stability of an originally stable vehicle can be decreased by deploying the fins; increasing the maneuverability, similar to tuna's dorsal fins. As shown in Figure \ref{fig:morphing}A, the morphing fins were usually retracted during straight runs in order to increase the stability. When the vehicle is required to make a quick heading change, the morphing fins were deployed, as shown in \ref{fig:morphing}B, to destabilize the body, increasing the maneuverability. In addition, the morphing fins were able to be articulated, as shown in Figure \ref{fig:morphing}C, providing a turning moment, to further increase the turning rate. The theoretical derivations of stability-maneuverability criteria, and the mathematical representation of forward-located morphing fins were presented in-detail, in our prior work \cite{randeni2022bioinspired}; therefore, a concise summary is given here in Appendix \ref{sec:appendix:morphing_theory}.  

The morphing payload module was developed as an independent section, that can be outfitted to any place within the mid-body of the base vehicle. Our previous work \cite{randeni2022bioinspired} investigated the variation of the stability index with the location of morphing fins; concluding that a larger stability index variation can be achieved when the fins were located closer to the nose-tip of the vehicle. Thus, in both the \textit{Morpheus} and \textit{Perseus} vehicles, we placed the morphing fin payload module immediately after the nose-cone.

\subsection{Morphing fin hardware design}

The morphing fin hardware design, as shown in Figures \ref{fig:morphing}D - \ref{fig:morphing}F, consists of two morphing fins that were driven in and out of the hull through fin cutouts by push rods. The push rods and their mounting arms were in turn driven by a 32 pitch gearwheel and an oil-filled micro-servo. The fins and rods moved in unison, providing symmetric deployment and retraction. The range of the fin movement was such that when fully retracted, just a few millimeters of fin protrudes from the hull, while when fully deployed the fin bottom clears the hull, allowing for articulation.  

The deploying mechanism was mounted on a carriage with ability to swing approximately 20 degrees to either side, thereby resulting in fin articulation. A 3D-printed rack on the carriage was driven by a 32 pitch gearwheel and an oil-filled articulation micro-servo. Similar to deployment action, the articulation of the fins was also symmetric. 

The morphing mechanism was nested into a free-flood hull chamber. Watertight channels were designed on either sides of the chamber, providing watertight wiring channels to run electrical cables across the morphing fin module.   

\subsection{Morphing fin software design} \label{sec:morphing_sw}

As discussed in Section \ref{sec:pactuatormap} and illustrated in Figure \ref{fig:low-level_control}, the platform-independent control correctives produced by the {\tt pControlEngine} were converted to actuator commands of the \textit{Morpheus} vehicle in {\tt pActuatorMap\_morpheus}. The adaptive morphing argument was also embedded within this application.
 
The morphing fins were controlled according to the magnitude of the heading error. The fins were deployed if the heading error (i.e. the difference between the desired and current vehicle heading) is larger than 30$^{\circ}$. Once deployed, the fins were actively controlled with an equal but opposite angle to the rudder deflection. When the heading error reduced to less than 5$^{\circ}$, the morphing fins were retracted.

% ==============================================
\section{Results and discussion}
% ==============================================

\begin{figure}[h]
\centering
\includegraphics[trim=110 0 110 0, clip,  width=0.48 \textwidth]{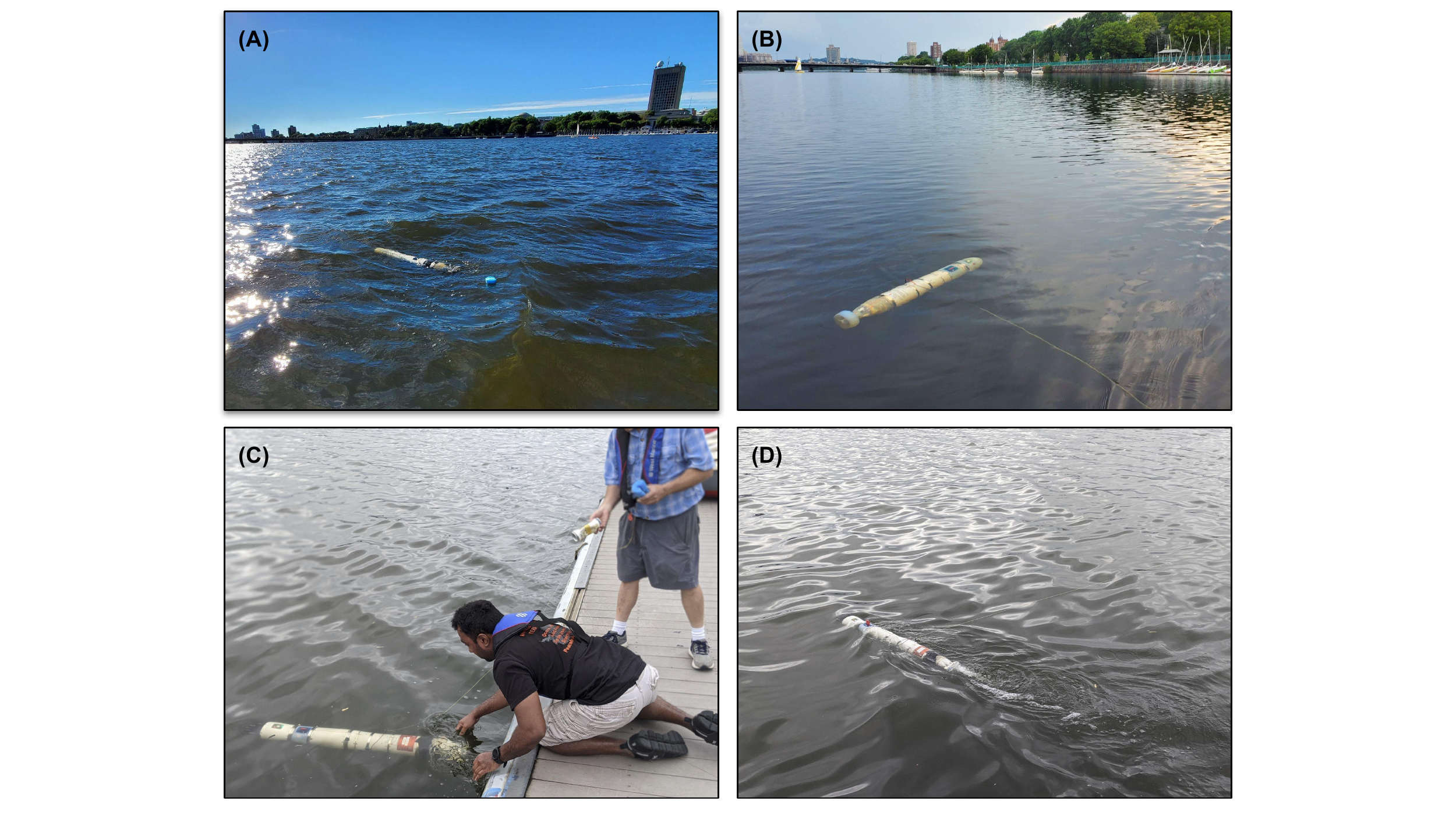}
\caption{In-water deployments were conducted in the Charles river, Massachusetts, USA, adjacent to the MIT Sailing Pavilion. Field deployment photos of (A) \textit{MIT-EMATT} AUV, (B) \textit{Morpheus} AUV, and (C-D) \textit{Perseus} AUV.}
\label{fig:field_photos}
\end{figure}

The original \textit{MIT-EMATT} base vehicle, the optimized base vehicle and \textit{Morpheus} AUV were all extensively field tested in-water in the Charles river, Massachusetts, USA by conducting hundreds of hours of operations over a period of two years (see Figure \ref{fig:field_photos}). In this section, we present in-water test results from a set of randomly picked missions. 

\begin{figure}[h]
\centering
\includegraphics[trim=0 0 0 0, clip,  width=0.48 \textwidth]{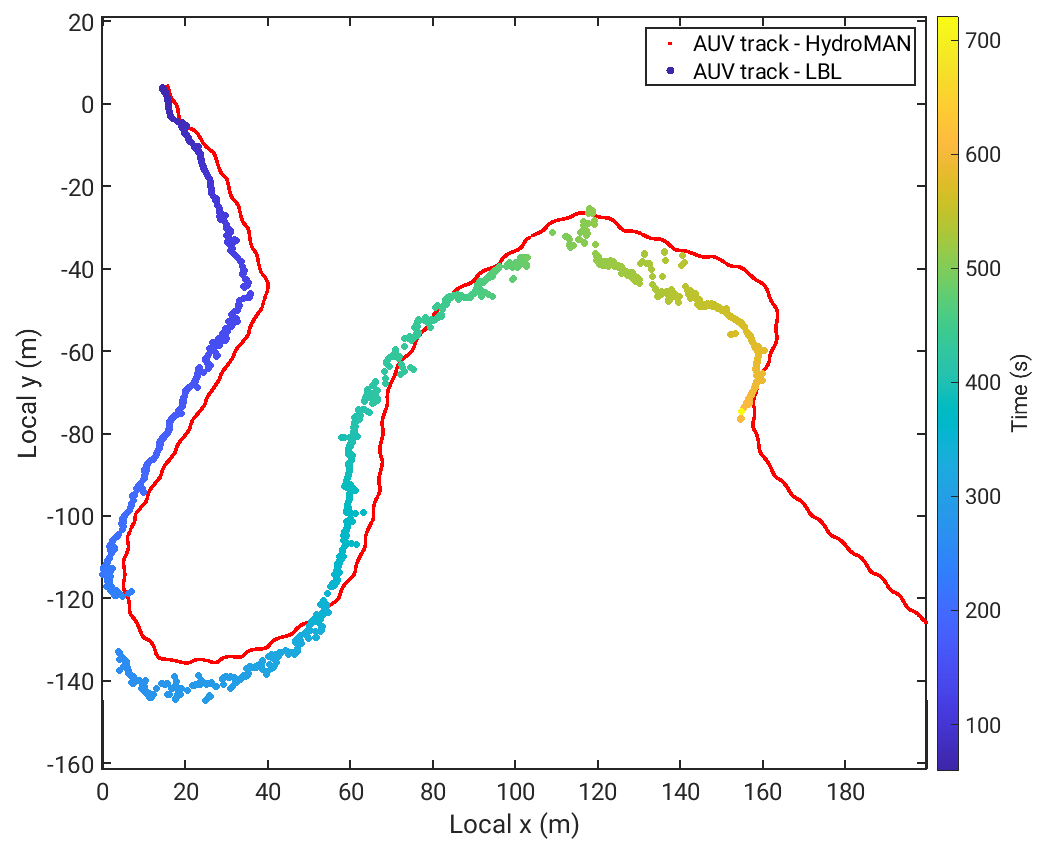}
\caption{Comparison of the HydroMAN navigation solution, which was limited to the depth sensor, IMU and the embedded vehicle flight dynamic model, against the navigation solution obtained from an LBL system.}
\label{fig:hydroman_nav}
\end{figure}

The vehicle tracks shown in this section were produced using the HydroMAN navigation solution. The base vehicles and \textit{Morpheus} AUV were limited to a depth sensor and an IMU. Therefore, the HydroMAN navigation engine was heavily relying on its embedded vehicle flight dynamic model. The HydroMAN navigation engine requires an initial vehicle motion response dataset to identify the parameters of the vehicle flight dynamic model \cite{randeni_hydroman,randeniAUV18}. In this work, we used the \textit{Perseus} vehicle configuration (shown in Figure \ref{fig:vehicle}E), which was outfitted with the piUSBL payload, to obtain the parameter estimation dataset. The piUSBL system was configured as a long baseline (LBL) system, and followed the same methodology as \cite{randeniAUV18} to estimate the vehicle flight dynamic model. Figure \ref{fig:hydroman_nav} compares the HydroMAN navigation solution against the LBL-based navigation solution, for validation and verification purposes.

\subsection{In-water tests of the \textit{MIT-EMATT} base vehicle}

\begin{figure}[h]
\centering
\includegraphics[trim=0 0 0 0, clip,  width=0.48 \textwidth]{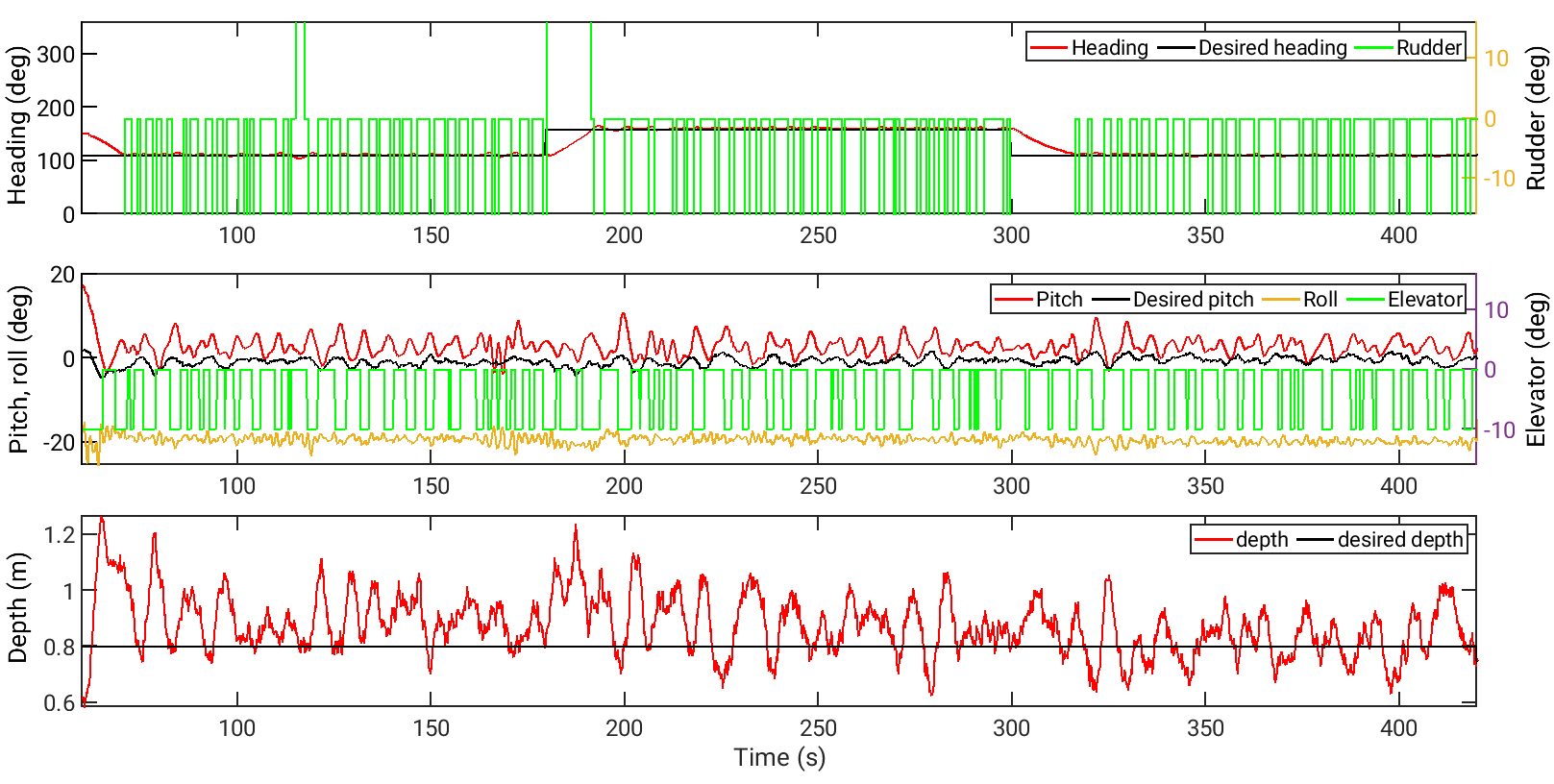}
\caption{A control response plot from one of the PID tuning runs conducted with the original \textit{MIT-EMATT} base vehicle, with a solenoid-driven, bang-bang controlled rudder and elevator. The top subplot shows the desired and actual heading responses of the vehicle together with bang-bang rudder commands. Middle plot shows the desired pitch (i.e. the output produced by the depth control loop) and actual pitch responses with bang-bang elevator commands. The roll response is also shown. The bottom plot illustrates the desired and actual depth responses.}
\label{fig:ematt_control}
\end{figure}

Figure \ref{fig:ematt_control} shows a typical control response plot of the original \textit{MIT-EMATT} base vehicle from an example zig-zag mission. The top subplot shows the desired and actual heading responses of the vehicle together with the rudder commands. As discussed in Section \ref{sec:tailcone}, the original \textit{MIT-EMATT} base vehicle tail-cone had a solenoid-driven rudder and elevator that only allowed bang-bang control. Therefore, as seen from Figure \ref{fig:ematt_control} top subplot, the rudder commands had only three positions; i.e. hard-to-port, hard-to-starboard and neutral. This resulted in around 5-10 degree amplitude oscillations in the heading response. 

Figure \ref{fig:ematt_control} middle subplot shows the desired and actual pitch responses of the vehicle, together with bang-bang elevator commands. As discussed in Section \ref{sec:controlengine}, the desired pitch was computed by the depth control loop within {\tt pControlEngine}; attempting to maintain the vehicle depth at the desired depth command. 
A constant roll angle of around 20 degrees was generally observed; primarily as a result of propeller torque. The original \textit{MIT-EMATT} did not have split rudders or split elevators that allowed implementation of active roll control. We addressed this drawback in the optimized vehicle by having individually controlled split rudders and split elevators. In addition, we also included fixed fins with a 3-degree constant angle of attack to counteract the rolling effect due to propeller torque. 

Figure \ref{fig:ematt_control} bottom subplot illustrates the desired and actual depth responses of the vehicle, which had an oscillation of around 0.2 - 0.4 m amplitude. This is primarily due to the band-bang control strategy and external disturbances.  

\begin{figure}[h]
\centering
\includegraphics[trim=0 0 0 0, clip,  width=0.48 \textwidth]{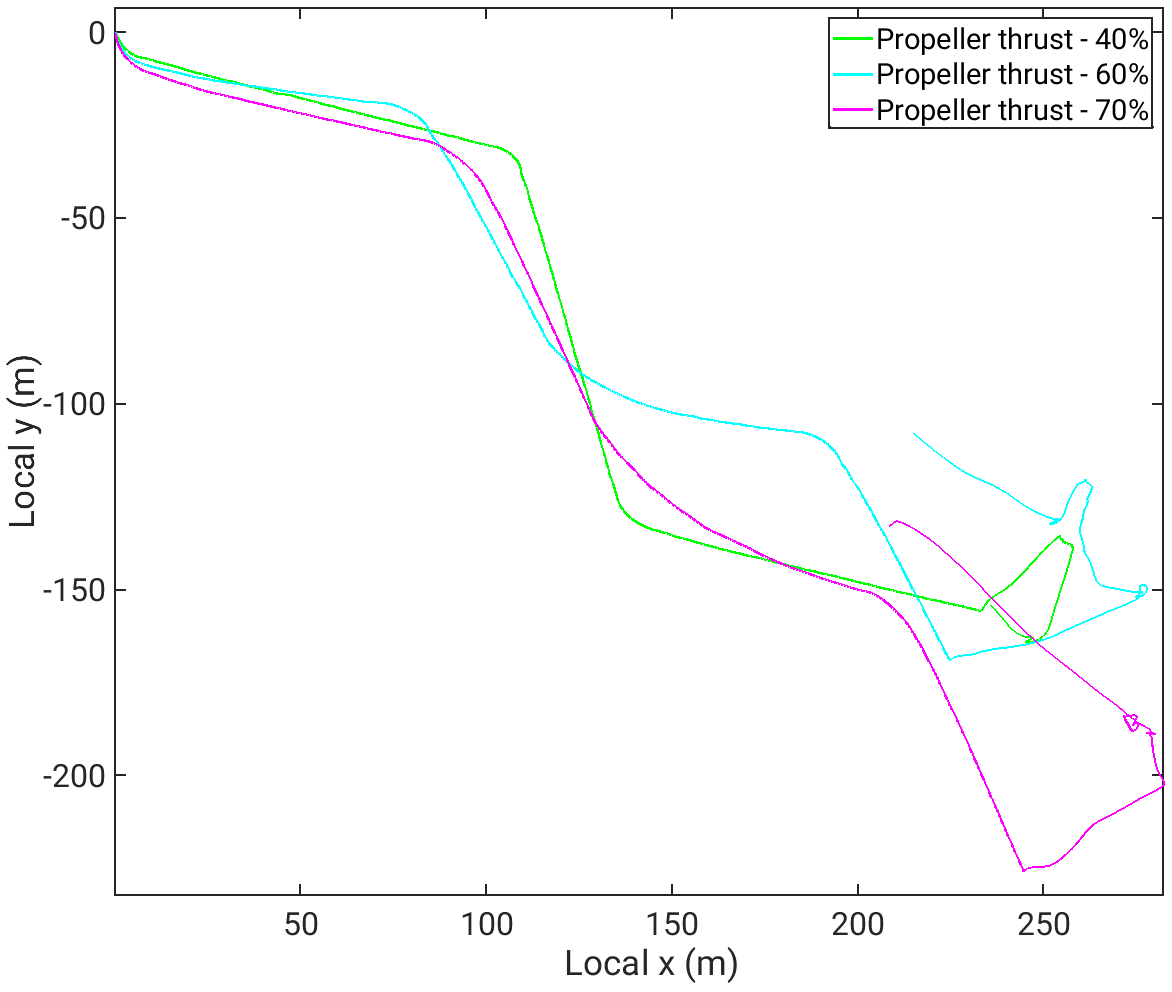}
\caption{Tracks of the original \textit{MIT-EMATT} base vehicle conducting three identical zig-zag pattern missions using {\tt pHelmPassive} at various propeller thrust percentages.}
\label{fig:ematt_xy}
\end{figure}

Figure \ref{fig:ematt_xy} shows the vehicle navigation tracks of the original \textit{MIT-EMATT} base vehicle from three identical zig-zag missions, conducted at different thrust percentage values for PID tuning. These missions were conducted using {\tt pHelmPassive}, which publishes pre-scripted, time triggered desired heading and desired depth commands. Such simplified missions were used during PID tuning and heading performance evaluation stages.  

\subsection{In-water tests of the \textit{Morpheus} AUV}

\begin{figure}[h]
\centering
\includegraphics[trim=0 0 0 0, clip,  width=0.48 \textwidth]{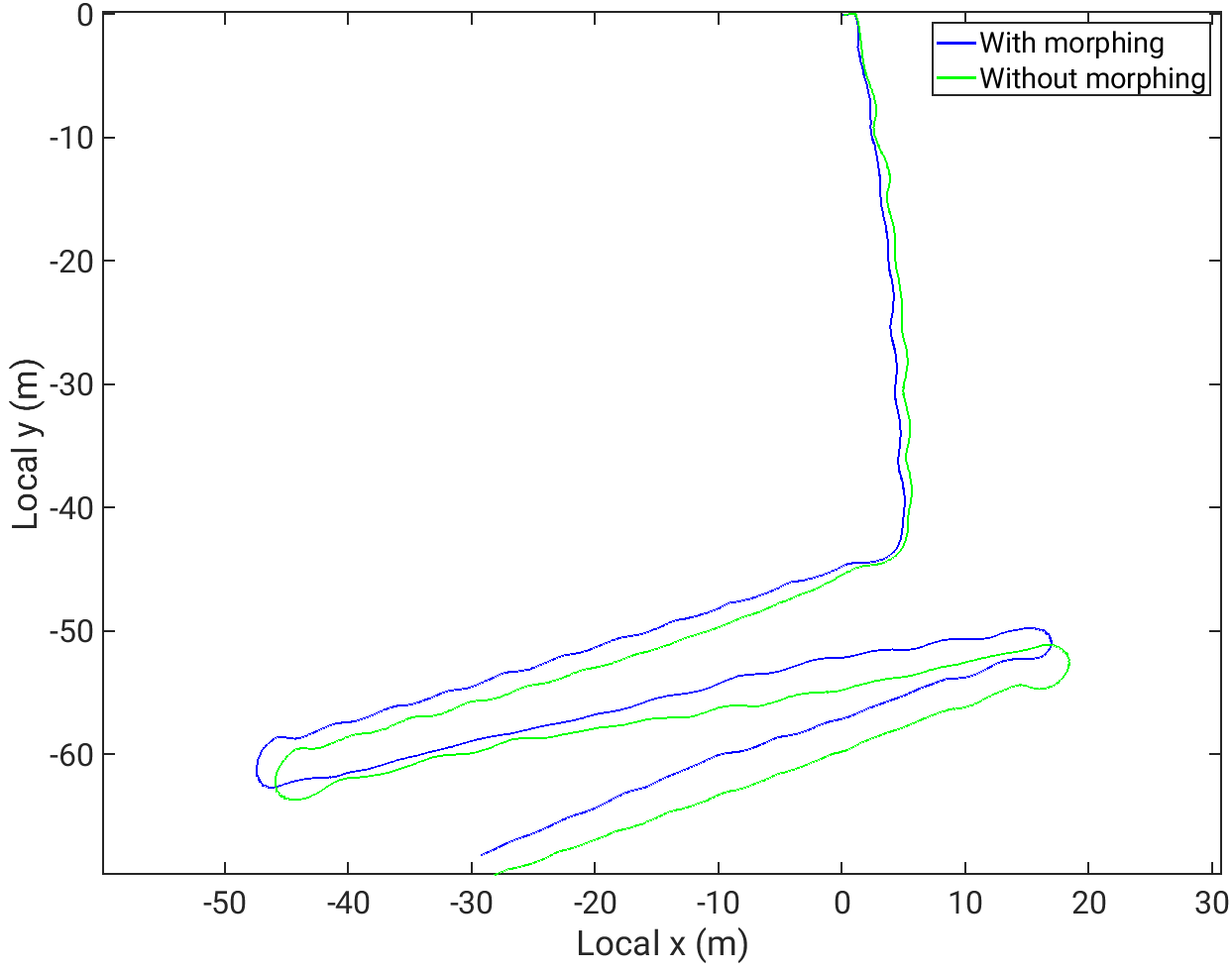}
\caption{Tracks of the \textit{Morpheus} AUV conducting two identical zig-zag missions using {\tt pHelmPassive}, with and without employing morphing fins.}
\label{fig:morpheus_xy}
\end{figure}

Similar to the \textit{MIT-EMATT} base vehicle, both the optimized base vehicle and \textit{Morpheus} AUV were intensively tested in-water for PID tuning and maneuverability-stability evaluations. Figure \ref{fig:morpheus_xy} shows the vehicle navigation track of the \textit{Morpheus} AUV for two identical zig-zag missions; one with morphing fins engaging according to the argument discussed in Section \ref{sec:morphing_sw}, and the second without engaging morphing fins. As seen, both runs provide small turning radii, with the run that engaged morphing fins outperforming the other. 

\begin{figure}[h]
\centering
\includegraphics[trim=0 0 0 0, clip,  width=0.5 \textwidth]{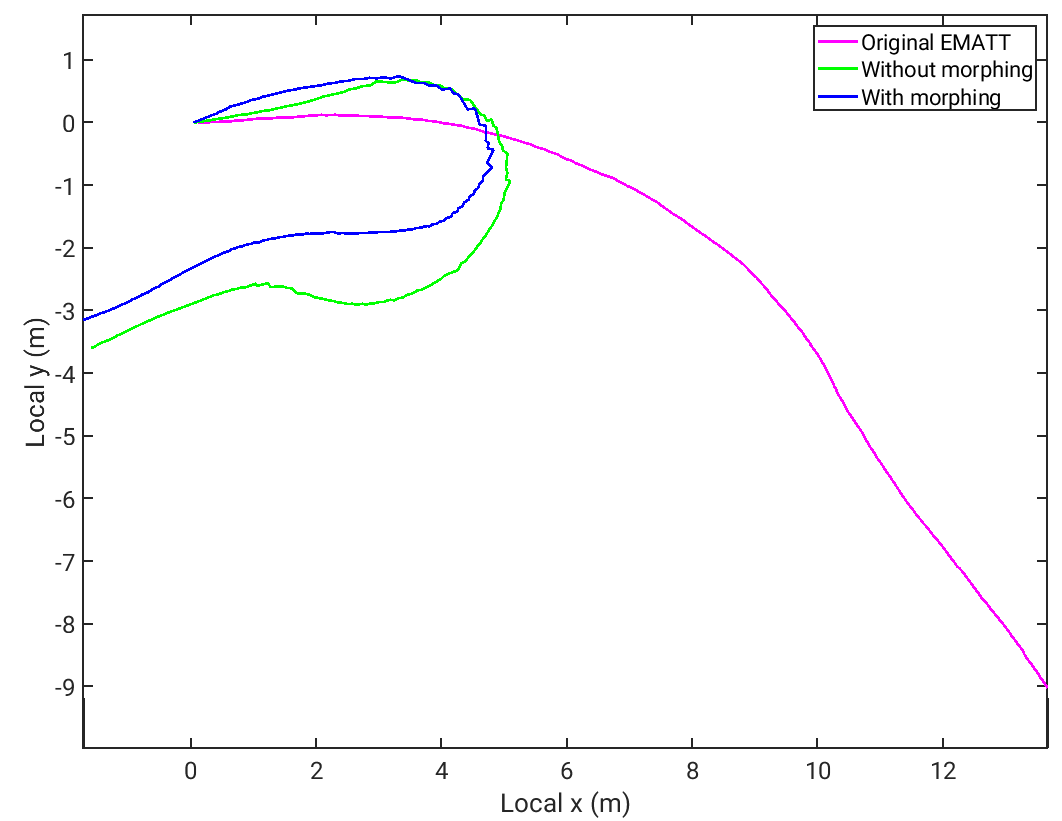}
\caption{A visual comparison of the turning radii between the original \textit{MIT-EMATT} vehicle (i.e around 10 m), \textit{Morpheus} without engaging morphing fins (i.e around 2.5 m), and \textit{Morpheus} with morphing fins (i.e around 1.5 m).}
\label{fig:turning_radius}
\end{figure}

Figure \ref{fig:turning_radius} illustrates a comparison of the starboard turns of the same runs. In addition, it also includes a similar turn of the original \textit{MIT-EMATT} base vehicle. The \textit{Morpheus} vehicle provided a turning radius of around 2.5 m when morphing fins were not engaged. The morphing fins were able to further reduce the turning radius down to approximately 1.5 m. In comparison, the turning radius of the original \textit{MIT-EMATT} vehicle was limited to around 10 m. As seen, a significant turning rate improvement was obtained through the use of morphing fins.

\begin{figure}[h]
\centering
\includegraphics[trim=0 0 0 0, clip,  width=0.48 \textwidth]{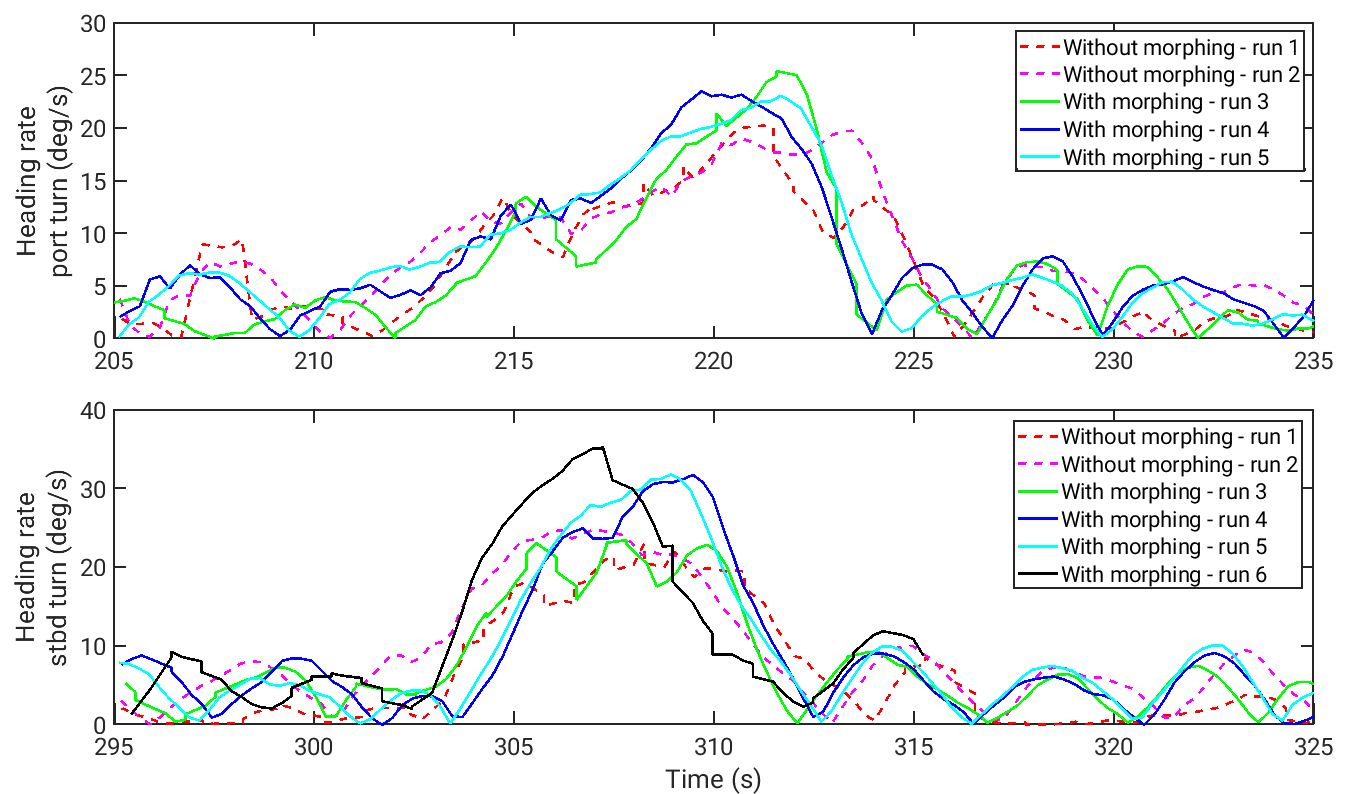}
\caption{The heading rate responses of the \textit{Morpheus} AUV observed during a set of randomly picked (upper) starboard and (lower) port turns; with and without engaging morphing fins.}
\label{fig:turn_rate}
\end{figure}

Figure \ref{fig:turn_rate} shows the turning rate responses of the \textit{Morpheus} AUV for six different example runs, both with and without engaging morphing fins. When comparing top and bottom subplots, the starboard turns always had a significantly higher turning rate as compared to port turn (i.e. approximately 10 deg s$^{-1}$  higher). We believe that this was as a result of the propeller torque favoring the starboard turns.    

As seen from Figure \ref{fig:turn_rate}, the \textit{Morpheus} AUV was able to showcase an exceptional turning rate of around 25-35 deg s$^{-1}$. A maximum turn rate improvement of around 35\% - 50\% was gained through the use of morphing fins.   

% ==============================================
\section{Conclusions}
% ==============================================

We designed and constructed an A-sized base AUV, augmented with a stack of modular and extendable hardware and software, including navigation, autonomy, control and high fidelity simulation capabilities. The base vehicle developed in this work was a derivation of the EMATT vehicle hullform, designed and produced by Lockheed Martin Corporation. During the first iteration, we used the original EMATT shell, including the original nose-cone, main-motor bay, and a free-flood tail-cone with solenoid-driven control surfaces; augmented with our own electronics and software stacks. In the second iteration of the base-vehicle, we hydrodynamically optimized nose and tail cones. The optimized nose-cone included an embedded GPS antenna, LED strobes, external pressure sensor and vacuum port; and the optimized tail-cone included four individually controlled, servo-based control surfaces. 

Subsequently, we extended the optimized base vehicle with a novel tuna-inspired morphing fin payload module (referred to as the Morpheus AUV), to achieve good directional stability and exceptional maneuverability; properties that are highly desirable for rigid hull AUVs, but are presently difficult to achieve because they impose contradictory requirements. The morphing fin payload allows the base AUV to dynamically change its stability-maneuverability qualities by using morphing fins, which can be deployed, deflected and retracted, as needed.

The original \textit{MIT-EMATT} base vehicle, the optimized base vehicle and \textit{Morpheus} AUV were all extensively field tested in-water in the Charles river, Massachusetts, USA by conducting hundreds of hours of operations over a period of two years. The \textit{Morpheus} vehicle  provided a turning radius of around 2.5 m when morphing fins were not engaged. The morphing fins were able to further reduce the turning radius down to approximately 1.5 m. In comparison, the turning radius of the original \textit{MIT-EMATT} vehicle was limited to around 10 m. The \textit{Morpheus} AUV was able to showcase an exceptional turning rate of around 25-35 deg s$^{-1}$. A maximum turn rate improvement of around 35\% - 50\% was gained through the use of morphing fins.

\section*{ACKNOWLEDGMENT}

This work was funded by a grant from Lockheed Martin Corporation; equipment were provided by the Naval Undersea Warfare Center (NUWC).

We greatly acknowledge Sekhar Tangirala, Stephen Bethel Jr., and Philip Nicolescu at Lockheed Martin Corporation, and Russell Sylvia (now at NUWC) for their technical discussions and continued support for this work. We also acknowledge Prof. Henrik Schmidt at MIT for his autonomy mission manager concept that was adapted to our system; and developers of open source libraries that were used in this work: Dr. Toby Schneider at gobysoft for NETSIM TCP library; Pierce Nichols for pBBBlue MOOS application; and Strawson Design for the BeagleBone Robotics Cape and robot control library. Many thanks to Emily Mellin, Nikolai Gershfeld, Suparnamaaya Prasad, Tyler Paine and Mike DeFilippo at MIT Sea Grant AUV Laboratory, Dr. Nick Rypkema at WHOI and MIT Sailing Pavilion staff for their support in field trials.

% ==============================================
%  APPENDIX
% ==============================================
\renewcommand{\thesection}{\Alph{section}}\setcounter{section}{0}

\section{Appendix -- Mathematical representation of morphing fins} \label{sec:appendix:morphing_theory}

\begin{figure}[h]
\centering
\includegraphics[trim=50 75 50 50, clip,  width=0.3 \textwidth]{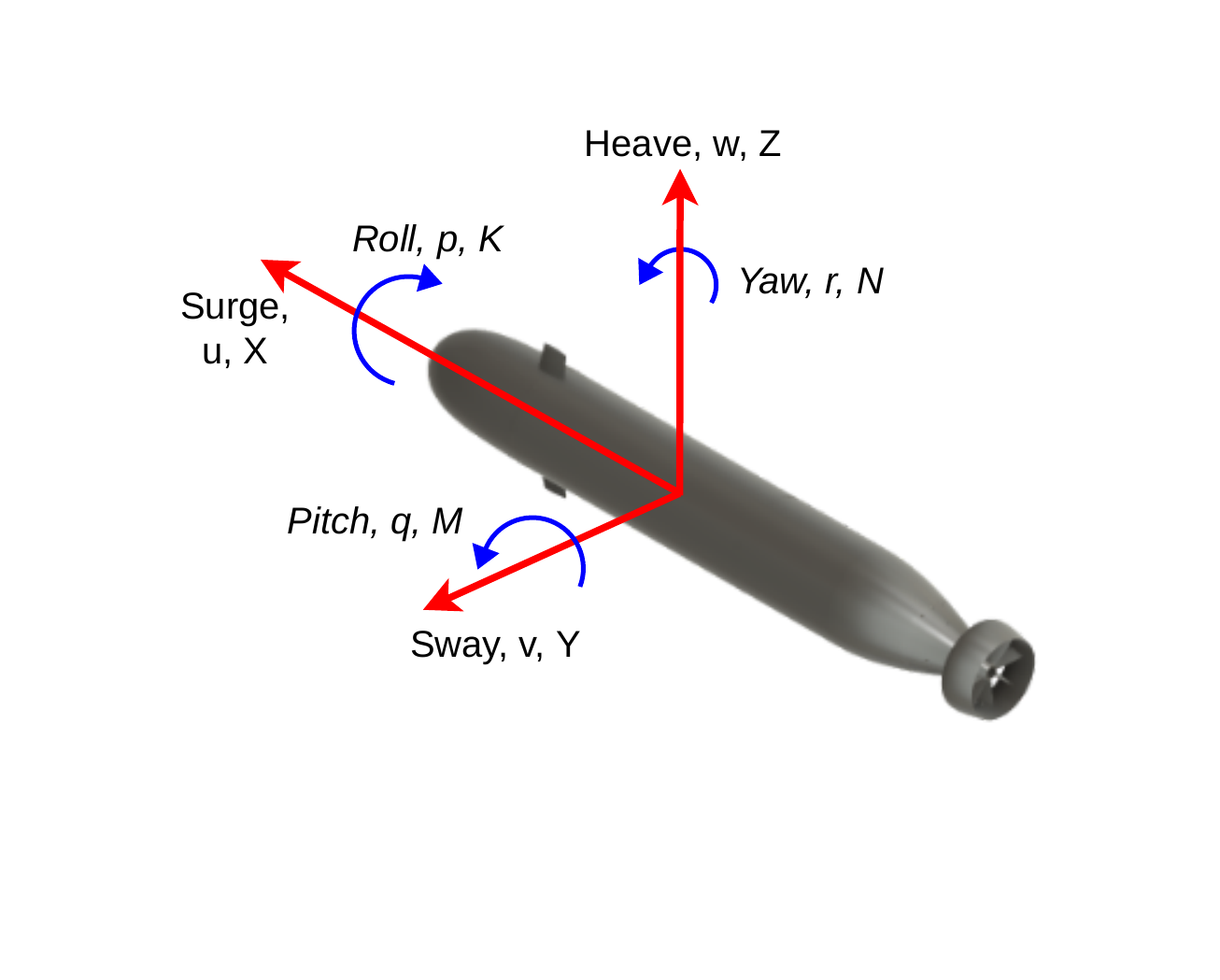}
\caption{The body-fixed reference frame used in this article. $X$, $Y$, $Z$ and $K$, $M$, $N$ are the body-fixed forces and moments along/around the surge, sway and heave axes of the AUV, respectively. The linear and angular velocities along/around the surge, sway and heave axes are \textit{u}, \textit{v}, \textit{w} and \textit{p}, \textit{q}, \textit{r}.}
\label{fig_axis_system}
\end{figure}

Utilizing the hydrodynamic coefficient representation of vehicle motion,
Triantafyllou et al. \cite{triantafyllou2020biomimetic} provides the theoretical derivation of the stability criterion for underwater vehicles, and how the stability-maneuverability is affected by the stern control surfaces and forward morphing fins. In this section, we summarize the theory given in Triantafyllou et al. \cite{triantafyllou2020biomimetic}, and further extend the derivation to show how the stability-maneuverability is affected by the size of stern control surfaces, forward morphing fins, and shroud. The body-fixed axes system and notations utilized in this article is shown in Figure \ref{fig_axis_system}.

\subsection{Directional stability versus maneuverability} \label{sec_stability_vs_maneuv}

Utilizing the hydrodynamic coefficient representation of vehicle motion \cite{maneuvering}, the linearized equations of sway and yaw motion of a torpedo-shaped vehicle, decoupled from surge, heave, roll and pitch motion, can be written as given in Equations \ref{eq_sway} and \ref{eq_yaw}.

\begin{equation}
Y = (m-Y_{\dot{v}})\dot{v} + (m x_{G} - Y_{\dot{r}})\dot{r} - Y_{v}v + (m U - Y_{r})r
\label{eq_sway}
\end{equation}

\begin{equation}
N = (I_{zz} - N_{\dot{r}})\dot{r} + (m x_{G} - N_{\dot{v}})\dot{v} - N_{v}v + (m x_{G} U - N_{r})r
\label{eq_yaw}
\end{equation}

\noindent where, $m$ is the mass, $I_{zz}$ is the moment of inertia about the origin, $x_{G}$ is the longitudinal location of the center of gravity, $U$ is the forward speed of the vehicle, $v$ and $\dot{v}$ are the sway velocity and acceleration, $r$ and $\dot{r}$ are the yaw velocity and acceleration. The hydrodynamic coefficients $-Y_{\dot{v}}$ and $-Y_{\dot{r}}$ denote the added mass in sway due to swaying and yawing acceleration, respectively. $-N_{\dot{v}}$ and $-N_{\dot{r}}$ denote the added moment of inertia due to sway and yaw acceleration. $Y_v$ and $Y_r$ are the linear resistance force in sway due to sway and yaw velocities, and $N_v$ and $N_r$ are the linear resistance moments in yaw due to sway and yaw velocities.

When the rudder is deflected to an angle $\delta$, after the transients die down and a steady turning at forward velocity U, yaw rate r, and side velocity v is achieved, the acceleration terms can be dropped. Then, Equations \ref{eq_sway} and \ref{eq_yaw} become Equations \ref{eq_sway_steady} and \ref{eq_yaw_steady}, respectively:

\begin{equation}
 - Y_{v}v + (m U - Y_{r})r = Y_{\delta} \delta
\label{eq_sway_steady}
\end{equation}

\begin{equation}
- N_{v}v + (m x_{G} U - N_{r})r = N_{\delta}\delta
\label{eq_yaw_steady}
\end{equation}

\noindent where, $Y_{\delta}$ and $N_{\delta}$ are the linear hydrodynamic coefficients of the rudder. Note that the rudder forces are taken to be a linear function of the rudder angle within this section.

Thus, the yaw rate, $r$ can be written as:

\begin{equation}
r = \frac{Y_{\delta} \delta}{C} (N_{v} Y_{\delta} - Y_{v} N_{\delta})
\label{eq_yawrate}
\end{equation}

\noindent where, the denominator $C$ can be shown to be the \textit{dynamic stability index}, $C$, as given in Equation \ref{eq_C} \cite{maneuvering}.

\begin{equation}
C = - Y_{v} (m x_{G} U - N_{r}) + N_{v} (m U - Y_{r})
\label{eq_C}
\end{equation}

If $C > 0$, the body is directionally stable, otherwise, it is linearly unstable. Equation \ref{eq_C} can be recasted as:

\begin{equation}
C = - Y_{v} (m U - Y_{r}) (x_{r} - x_{AC})
\label{eq_C_recast}
\end{equation}

\noindent where,

\begin{equation}
x_{r} = \frac{m x_{G} U - N_{r}}{m U - Y_{r}}
\label{eq_C_xr}
\end{equation}

\begin{equation}
x_{AC} = \frac{N_{v}}{Y_{v}}
\label{eq_C_ac}
\end{equation}

\noindent $x_{r}$ is the distance of the \textit{Center of Rotational motion} (CR) from the origin, i.e. the location where the side
force acts when the body performs a pure rotation at constant speed $U$ and (small) angular velocity $r$, and
$v = 0$. $x_{AC}$ is the distance from the origin to the \textit{Aerodynamic Center} (AC), i.e. the location where the side force acts when the body performs a steady translation at forward velocity $U$ and side velocity $v$, while $r = 0$ (what is referred to, also, as sideslip velocity).

As noted by \cite{triantafyllou2020biomimetic} \cite{triantafyllou2020biomimetic}, the aerodynamic center is a critical quantity in determining the body stability. Since $Y_{v}$ is always a negative quantity \cite{maneuvering}, and $(m U - Y_{r})>0$, as $m U$ is a large positive quantity, the stability criterion can be recast as:

\begin{equation}
x_{r} > x_{AC}
\label{eq_stability}
\end{equation}
As the difference between these two values increases, the linear stability of the vehicle increases while the maneuverability decreases as shown from equation (\ref{eq_yawrate}). 

\subsection{Presence of stern control surfaces} \hfill

The effects of the rudder are next added to the equations of motion, with the subscript \textit{b} corresponding to the bare body coefficients. Following \cite{triantafyllou2020biomimetic} \cite{triantafyllou2020biomimetic}, the updated hydrodynamic coefficients are:
\begin{equation} \label{eq:Y_bare+rudder}
    Y_v = Y_{v,b} + \frac{Y_\delta}{U}
\end{equation}
\begin{equation} 
    Y_r = Y_{r,b} - x_R\frac{Y_\delta}{U}
\end{equation}
\begin{equation} 
    N_v = N_{v,b} - x_R\frac{Y_\delta}{U}
\end{equation}
\begin{equation} \label{eq:N_bare+rudder}
    N_r = N_{r,b} - x_R^2\frac{Y_\delta}{U}
\end{equation}
where \textit{$x_R = N_\delta/Y_\delta$} is the location where the force acts on the rudder and $Y_\delta$ and $N_\delta$ are defined earlier. 

The stability index is now updated to take into account the effect of the rudder. Plugging the updated hydrodynamic coefficients into Equation (\ref{eq_C}) and denoting the stability index of just the bare body by \textit{$C_b$} leads to the following updated stability index: 
\begin{equation}
    C = C_b - A(mx_GU - N_{r,b} - Y_{v,b}\xi^2 -N_{v,b}\xi + (mU-Y_{r,b})\xi)
\end{equation}
which reduces to:
\begin{equation} \label{eq:C_updated}
    C = C_b - A[-Y_{v,b}\xi(x_{AC,b} + \xi) + (mU-Y_{r,b})(x_{r,b}+\xi)]
\end{equation}
where 
\begin{equation}
    A = \frac{Y_\delta}{U} = \frac{L_{rudder}}{U} < 0
\end{equation}
\begin{equation}
    \xi = -x_R > 0
\end{equation}

The yaw rate from Equation (\ref{eq_yawrate}) is calculated within linear theory as:
\begin{equation} \label{eq:final_yaw_rate}
    \frac{r}{\delta} = \frac{A}{CU}Y_{v,b}(x_{AC,b} + \xi)
\end{equation}

\subsection{The size of stern control surfaces} \hfill

 The lift generated by the stern control surfaces contributes to the hydrodynamic coefficients, as shown in Equations \ref{eq:Y_bare+rudder}-\ref{eq:N_bare+rudder}. To estimate how much lift is generated by the rudder, the general form for lift is used:

\begin{equation} \label{eq:rudder_lift}
    L_{rudder} = \frac{1}{2}{\rho}C_L(2S_{rudder})U^2
\end{equation}
where $C_L$ is the coefficient of lift and $S_{rudder}$ is the area of one of the two rudders. 

As \cite{triantafyllou2020biomimetic} \cite{triantafyllou2020biomimetic} concluded, the addition of these stern control surfaces can stabilize an initially unstable vehicle, as long as it is above a threshold value that provides stability. Rearranging Equation \ref{eq:C_updated} determines what this threshold value for \textit{A} should be in order to bring the stability index $C$ from a negative to a positive value. Since \textit{A} is determined by the amount of lift generated for a certain speed, the size of the rudder is what keeps this value close to the threshold value. If the size of the rudder increases, resulting in an increase in \textit{A}, the vehicle surpasses the stability threshold, becoming more stable and reducing the turning rate of the vehicle. The value of \textit{C} should remain close to this stability transition in order for the rudder to have a significant effect on the turning rate. 

\subsection{The presence of forward morphing fins} \label{sec_theory_dorsal_presence} \hfill

The addition of forward morphing fins to an underwater vehicle has also an effect on the stability and maneuverability of the vehicle. The hydrodynamic coefficients from Equations \ref{eq:Y_bare+rudder}-\ref{eq:N_bare+rudder} are updated to take into account the forward morphing fin following a similar process used for the rudder \cite{triantafyllou2020biomimetic}: 
\begin{equation} \label{eq:fin_formula_1}
    Y_v = Y_{v,b} + \frac{Y_\delta}{U} + \frac{Y_{f\delta}}{U}
\end{equation}
\begin{equation} 
    Y_r = Y_{r,b} - x_R\frac{Y_\delta}{U} - x_f\frac{Y_{f\delta}}{U}
\end{equation}
\begin{equation} 
    N_v = N_{v,b} - x_R\frac{Y_\delta}{U} -x_f\frac{Y_{f\delta}}{U}
\end{equation}
\begin{equation} \label{eq:fin_formula_2}
    N_r = N_{r,b} - x_R^2\frac{Y_\delta}{U} - x_f^2\frac{Y_{f\delta}}{U}
\end{equation}

\noindent where \textit{$x_f = N_{f\delta}/Y_{f\delta}$} is the location where the force acts on the vehicle, \textit{$Y_{f\delta}$} is the fin force per unit rudder angle and \textit{$N_{f\delta}$} is the moment per unit angle. Once again, the stability index in Equation \ref{eq:C_updated} is updated to reflect the effects of the forward morphing fin:

\begin{eqnarray} \label{eq:C_updated_fin}
    C = C_b - &A[(x_r+\xi)(mU-Y_{r,b})-Y_{v,b}\xi(x_{AC} + \xi)] \nonumber\\ &-B[(x_r-\eta) (mU-Y_{r,b})-Y_{v,b}\eta(\eta-x_{AC})] \nonumber\\  &+2AB(\eta^2+\xi^2)
\end{eqnarray}

\noindent with \textit{$B=\frac{Y_{f\delta}}{U} < 0$} and $\eta=x_f$. 

Hence, the yaw rate in Equation \ref{eq:final_yaw_rate} can be calculated with the updated stability index, \textit{C}, that takes into account the effects of both the rudder and forward morphing fin.  The values of $mU-Y_{r,b} > 0$ and $-Y_{v,b} >0$ require that $\eta > x_{r}$ and $\eta < x_{AC}$. In other words, the forward morphing fin needs to be positioned ahead of the bare body center of rotational motion and behind the bare body aerodynamic center of the vehicle, requiring that the bare body vehicle be initially directionally unstable. For a given stable vehicle and rudder configuration, the main way to increase the turning rate is to decrease the stability parameter. 

\subsection{The size and angle-of-attack of forward morphing fins} \label{sec_theory_dorsal_deflection} \hfill

A fish bends its body when it initiates a turn \cite{triantafyllou2020biomimetic}. Since the forward morphing fins are located ahead of the center of gravity, the fins deflect in the opposite direction of the rudder. This can be adapted to the coefficients derived so far for the case when $\delta_f = -\delta$, the magnitude of the forward morphing fin and rudder angles are equal but in opposite directions. The deflection does not affect the stability criterion. The turning rate for a vehicle with both rudder and forward morphing fins is:

\begin{eqnarray} 
\label{eq:r_full_equation}
    \frac{r}{\delta} = \frac{1}{CU}[(-A)(-Y_{v,b})(x_{AC} + \xi)+ \nonumber\\ (-B)(-Y_{v,b})(\eta-x_{AC}) \nonumber\\ + 2(-A)(-B)(\eta+\xi)]
\end{eqnarray}

\noindent where $\eta=x_f$ is the fin position, $\xi=-x_R$ is the rudder position, $A = \frac{Y_\delta}{U}$ and $B = \frac{Y_{f\delta}}{U}$. The last term in the equation contributes the strongest increase in the turning rate since $-A>0$, $-B>0$, and $-Y_{v.b}>0$ \cite{triantafyllou2020biomimetic}. In order to increase the rate of turning of the vehicle, the forward morphing fins should have a comparable size and lift generation as the rudder.

\bibliographystyle{IEEEtran}
\bibliography{main}

\end{document}